\documentclass[10pt,twocolumn,letterpaper]{article}

\usepackage{wacv}              

\usepackage{algorithm}
\usepackage{algpseudocode}
\usepackage{float}
\usepackage{placeins}
\usepackage{booktabs}
\usepackage{multirow}
\usepackage[T1]{fontenc}

\usepackage[pagebackref,breaklinks,colorlinks,allcolors=wacvblue]{hyperref}
\usepackage{xcolor}
\usepackage[accsupp]{axessibility}  

\definecolor{wacvblue}{rgb}{0.21,0.49,0.74}

\def\wacvPaperID{2414} 
\def\confName{WACV}
\def\confYear{2027}

\title{Beyond Visible Evidence: Open-World Amodal Counting Under Spatial Occlusion}

\author{
Safaeid Hossain Arib,
Rabeya Akter,
Abdul Monaf Chowdhury,\\
Md. Jubair Ahmed Sourov,
Md Mehedi Hasan\\
University of Dhaka\\
{\tt\small safaeid48@gmail.com,
rabeyaakter231023@gmail.com,
monafabdul15@gmail.com}\\
{\tt\small sourovjubair@du.ac.bd,
mmhasan@du.ac.bd}
}

\definecolor{bottlegreen}{RGB}{0,189,98}

\begin{document}
\maketitle

\begin{abstract}
  Object counting has achieved remarkable success on visible instances, yet existing open-world counting methods remain vulnerable under spatial occlusion. This limitation arises because backbone networks encode only visible surfaces and therefore lack the missing target evidence needed for amodal enumeration. To address this, we present \textbf{CountOCC}, an open-world amodal counting framework that reconstructs occlusion-aware features through hierarchical multimodal guidance. CountOCC integrates spatial context from visible regions with semantic guidance from text and visual embeddings to reconstruct features at occluded locations across multiple pyramid levels. We further enforce attention consistency between occluded and unoccluded views, encouraging spatially aligned target-conditioned attention maps. Together, these mechanisms preserve counting-relevant evidence needed for accurate amodal counting. For evaluation, we construct FSC-147-OCC and CARPK-OCC and further evaluate on CAPTURe-Real and KINS datasets. Compared with CountGD, CountOCC reduces MAE by 26.66\%/20.32\% on FSC-147-OCC validation/test, 47.52\% on CARPK-OCC, 28.52\% on CAPTURe-Real, and 66.23\% on KINS, demonstrating robust open-world amodal counting under spatial occlusion. Code: \href{https://github.com/Safaeid48/CountOCC}{https://github.com/Safaeid48/CountOCC}.
\end{abstract}
    
\section{Introduction}
\label{sec:introduction}

\begin{figure}[t]
  \centering
   \includegraphics[width=0.9\linewidth]{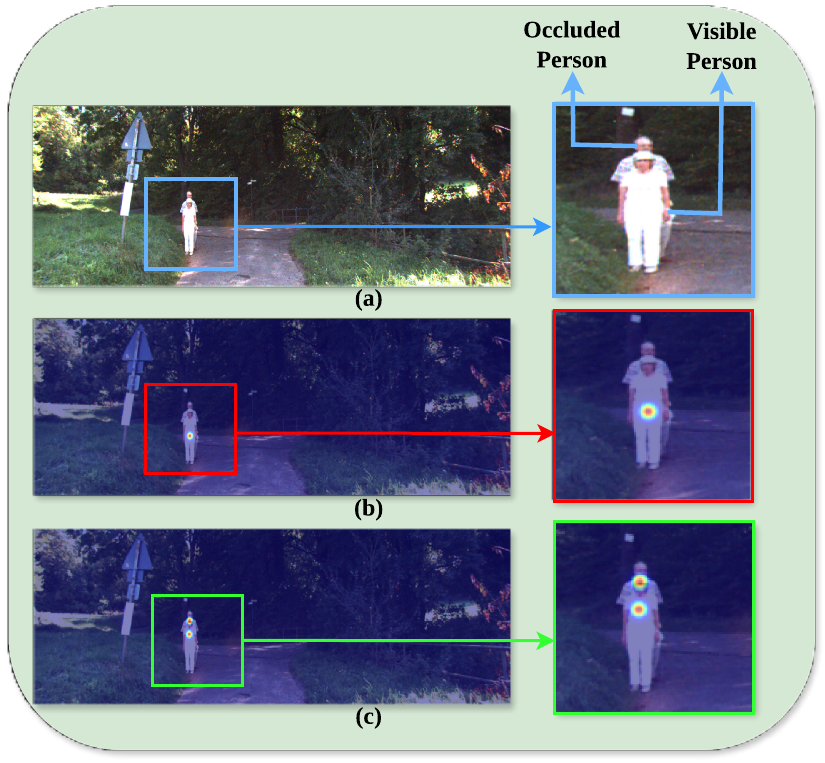}
   \caption{\textbf{Open-world amodal counting under spatial occlusion.} Example on an image from the KINS dataset: (a) the image contains two people, one visible and one occluded; (b) the CountGD architecture counts the visible person but misses the occluded person; (c) CountOCC counts both the visible and occluded persons, yielding the correct total count.}
   \label{fig:count-problem}
\end{figure}

Humans possess a remarkable ability to infer the existence of occluded objects from limited visual cues, an essential skill for navigating complex environments \cite{kanizsa1979organization, olson2004neuronal, otsuka2006development, wynn1990children, zhan2020self}. In practical scenarios such as parking lots, retail shelves, and agricultural fields, objects are often hidden by image-plane spatial occluders, such as foreground clutter, neighboring instances, or structural elements, but accurately estimating their total quantity remains crucial for inventory control, traffic management, and autonomous systems. In this work, we study open-world amodal counting under single-image spatial occlusion, where a 2D occlusion mask indicates regions of the image where target evidence may be missing. The task is to enumerate visible, partially visible, and occluded instances of arbitrary object categories specified at inference time, without requiring retraining for novel categories. This setting is distinct from multi-view or volumetric counting, where hidden counts depend on 3D geometry, depth, or physical occupancy. Although recent open-world counting methods demonstrate strong performance with fully visible objects, they systematically undercount when faced with spatial occlusion. 


Recent progress in open-world object counting has been driven by flexible input specifications. Methods such as CountGD~\cite{amini2024countgd}, LOCA~\cite{Dukic_2023_ICCV}, and CounTR~\cite{liu2022countr} allow users to define target categories through visual exemplars, text descriptions, or both, eliminating the need for class-specific retraining. However, these methods share a fundamental limitation, as they assume target objects remain predominantly visible. When faced with occlusion, they lack an explicit mechanism for amodal reasoning and instead count only what is directly observable, often missing hidden instances as illustrated in \cref{fig:count-problem}(b). 
This outcome follows naturally from their formulation, which does not explicitly model fully occluded objects when the goal is to estimate the total number of instances in the scene. 
The limitation is architectural, since direct feature extraction cannot recover target representations once foreground or structural occluding surfaces intervene in the image plane, leaving no mechanism to reconstruct the missing evidence.

We address this limitation by introducing CountOCC, an open-world amodal counting framework that reconstructs target-relevant features at occluded spatial locations, targeting the practically prevalent setting where foreground clutter or structural elements hide objects within a single image. Our key insight is that accurate amodal counting does not require photorealistic image completion; instead, it requires recovering amodal counting evidence in the feature space used by the counting model. At the core is a Feature Reconstruction Module (FRM) that operates across multiple hierarchical levels, integrating information from visible regions with semantic guidance from text-visual prompts to explicitly recover features at occluded locations. Rather than accepting corrupted encodings from occluding surfaces, FRM approximates the clean-image features the counting backbone would have produced at masked locations. We complement this with Visual Equivalence (VisEQ) supervision, which enforces consistency between occluded and unoccluded views through gradient-based attention alignment. This dual supervision at feature and attention levels ensures reconstructed representations preserve target-specific evidence needed for counting, enabling robust counting performance even under occlusion. 


Although the recent CAPTURe-Real benchmark~\cite{pothiraj2025capture} evaluates amodal counting through pattern completion in structured environments, it focuses primarily on regular arrangements where occluded instances follow predictable patterns. Models can infer hidden objects by extrapolating visible spatial regularities, but such pattern-based inference may be less reliable in unstructured natural scenes where objects exhibit irregular arrangements. To provide a comprehensive evaluation, we create occlusion-augmented versions of FSC-147~\cite{ranjan2021learning} and CARPK~\cite{hsieh2017drone} (FSC-147-OCC and CARPK-OCC), while preserving original splits and annotations. We further evaluate on CAPTURe-Real and KINS to assess performance under structured and naturally occurring real-world occlusions. Our method achieves substantial improvements across these benchmarks. 

These results establish state-of-the-art performance among the evaluated open-world counting methods under occlusion and provide a comprehensive evaluation framework for future research in this domain.

Our core contributions are summarized below:
\begin{itemize}
  \item We present CountOCC, an open-world amodal counting framework under single-image spatial occlusion that accurately quantifies target categories across both observable and occluded spatial regions.
  \item We introduce a Feature Reconstruction Module that reconstructs occlusion-aware feature representations at masked spatial locations, complemented by a Visual Equivalence objective that enforces attention consistency between occluded and unoccluded views.
  \item We constructed occlusion-augmented versions of the FSC-147 and CARPK datasets (FSC-147-OCC and CARPK-OCC) and evaluate CountOCC on these controlled benchmarks together with CAPTURe-Real and KINS datasets. 
  \item We provide substantial experimental analysis and ablation of CountOCC, and establish a new state-of-the-art for amodal object counting. 
\end{itemize}

\section{Related Work}
\label{sec:related_works}


\noindent \textbf{Open-world object counting.}
Object counting has progressed from class-specific detectors trained per category~\cite{abousamra2021localization, zhang2015cross, liu2019context, dai2019video, falk2019u, zavrtanik2020segmentation, ranjan2021learning} to open-world models~\cite{AminiNaieni23, kang2024vlcounter, liu2022countr, shi2022represent, you2023few} that adapt to novel target categories at test time. A key catalyst was FSC-147~\cite{ranjan2021learning}, which established few-shot, class-agnostic counting using a handful of visual exemplars. Early efforts span GMN~\cite{lu2018class}, which frames counting as feature matching between exemplar and image regions; FamNet~\cite{ranjan2021learning}, which introduces test-time adaptation via feature correlation; and SAFECount~\cite{you2023few}, which strengthens generalization through support-driven feature enhancement. Building on transformers, CounTR~\cite{liu2022countr} leverages cross-attention to fuse image and exemplar cues before regressing density maps, while LOCA~\cite{Dukic_2023_ICCV} iteratively adapts class prototypes with a learnable similarity metric. CountGD~\cite{amini2024countgd} builds on GroundingDINO~\cite{liu2024grounding} with learned cross-modal attention, enabling target specification through text prompts, visual exemplars, or both, while OmniCount~\cite{mondal2025omnicount} extends open-vocabulary counting to multiple queried categories using semantic and geometric priors. Yet across these lines of work, target instances are assumed to be fully or largely visible. These architectures do not explicitly model spatially occluded instances, causing hidden regions to be ignored or treated as non-target evidence in cluttered scenes.


\noindent \textbf{Multi-modal specification for counting.}
Open-world counters specify targets through visual exemplars, text prompts, or both. Visual exemplar-driven frameworks typically achieve superior accuracy through direct appearance matching, but they require user-provided bounding boxes at inference~\cite{Dukic_2023_ICCV, gong2022class, liu2022countr, lu2018class, nguyen2022few, ranjan2021learning, shi2022represent, yang2021class, you2023few, lin2022scale}. In contrast, recent text-only approaches eliminate this overhead by grounding natural language in visual features using pre-trained vision-language models~\cite{jiang2023clip, AminiNaieni23, dai2024referring, xu2023zero}. However, language alone can be less expressive for fine-grained appearance, texture, part, and scale cues, causing text-only methods to trail exemplar-based approaches on visually confusable categories. This accuracy-convenience trade-off has motivated hybrid designs such as DAVE~\cite{pelhan2024dave}, which employs separate modality pathways, whereas CountGD~\cite{amini2024countgd} unifies visual exemplars and textual prompts through learned cross-modal attention. Despite this flexibility, these frameworks still rely on observable target evidence. Under spatial occlusion, the image backbone only observes the visible surface at the masked location and cannot directly encode the missing evidence behind it; without explicit reconstruction at occluded locations, counting performance degrades.


\noindent \textbf{Amodal counting under occlusion.} 
Several counting domains address partial visibility, including crowd counting~\cite{zhou2024multi, wang2024dual, fan2023multi, liang2023crowdclip, babu2022completely, cho1999neural}, cell counting~\cite{bera2015partially, xie2018microscopy, flaccavento2011learning}, and crop yield estimation~\cite{wang2021occlusion}. These methods improve robustness to overlap and clutter but generally rely on at least partial target visibility. Amodal completion and segmentation instead infer hidden object structure: SeGAN~\cite{ehsani2018segan} predicts invisible regions, while ORCNN~\cite{follmann2019learning} and ASN~\cite{qi2019amodal} estimate complete amodal instance masks. Unlike CountOCC, these methods recover object extent rather than category-conditioned counts. Amodal counting with additional physical cues includes LiDAR-based retail counting~\cite{jenkins2023countnet3d} and multi-view 3D stacked-object counting~\cite{dumery2025counting}. CountOCC instead targets single-image open-world counting under 2D spatial occlusion. CAPTURe~\cite{pothiraj2025capture} is most related, but focuses on inferring hidden objects from regular spatial patterns, whereas real scenes may contain irregular layouts. Recent evaluations further show substantial degradation of vision-language models and open-world counters under occlusion~\cite{li2024naturalbench, wang2025seeing, qharabagh2024lvlm}. We address this gap through occlusion-aware feature reconstruction and clean--occluded attention consistency.
\section{Methodology}
\label{sec:methodology}

\begin{figure*}[t]
  \centering
   \includegraphics[width=0.8\linewidth]{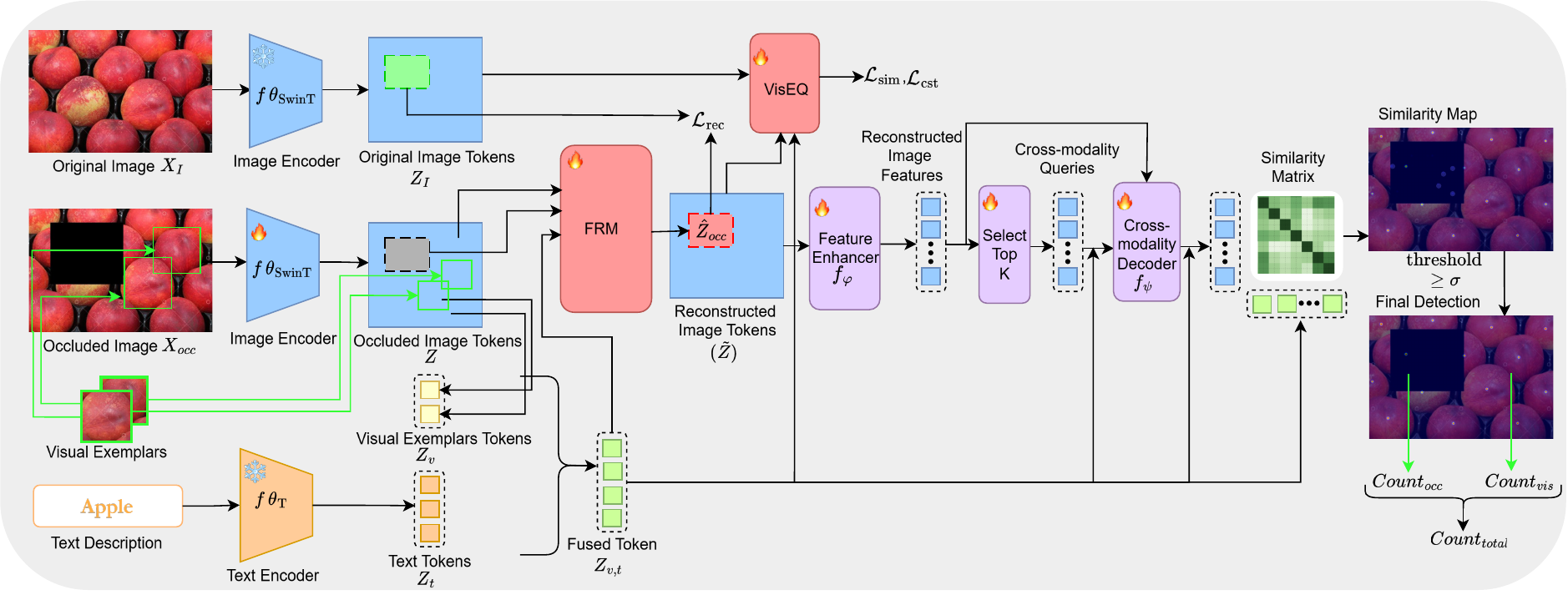}
\caption{\textbf{The CountOCC architecture.}
CountOCC extends CountGD with FRM for masked-region feature reconstruction and VisEQ for clean-occluded attention alignment, enabling amodal count estimation under occlusion.}
    \label{fig:methodology_overview}
\end{figure*}


\begin{figure*}[t]
\centering

\subcaptionbox{\textbf{Feature Reconstruction Module (FRM)}
\label{fig:feature-reconstruction-module}}%
[0.40\textwidth]{%
    \includegraphics[width=\linewidth]
    {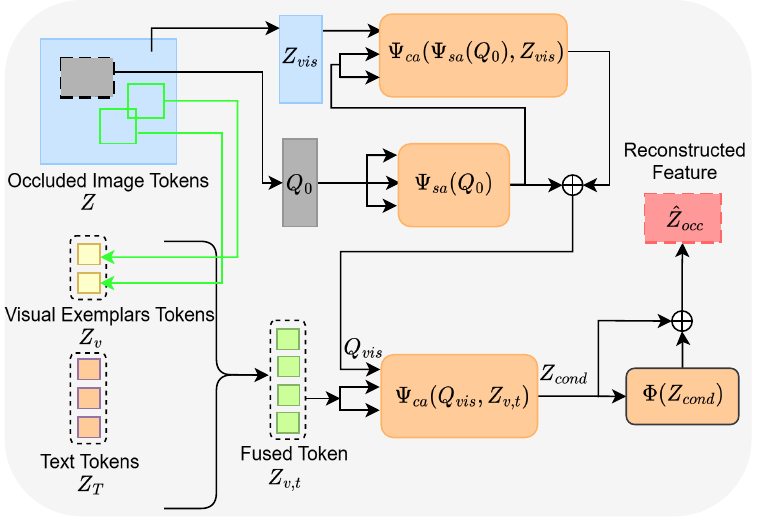}%
}
\hspace{0.035\textwidth}
\subcaptionbox{\textbf{Visual Equivalence (VisEQ) supervision}
\label{fig:visual-eq}}%
[0.40\textwidth]{%
    \includegraphics[width=\linewidth]
    {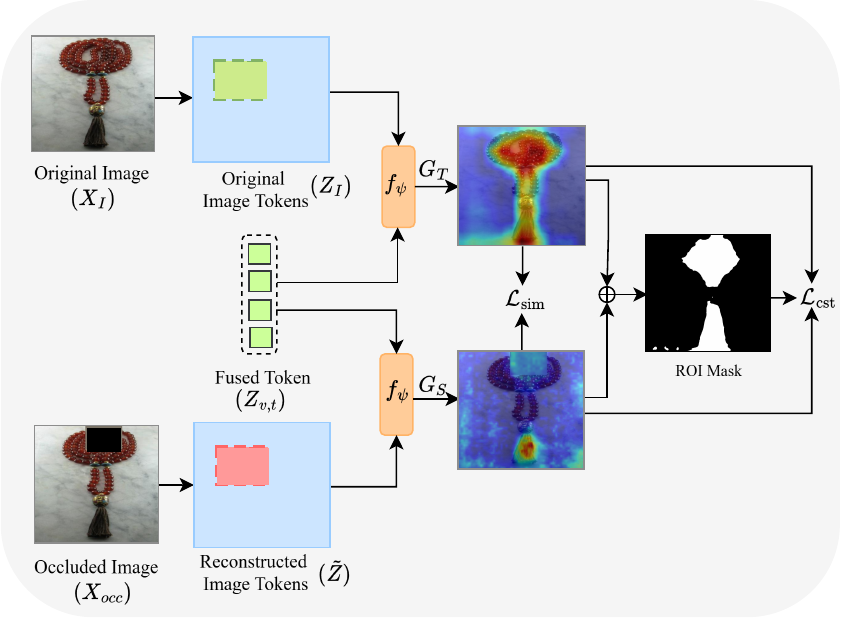}%
}

\caption{\textbf{CountOCC components.}
(a) FRM reconstructs prompt-conditioned features at occluded locations.
(b) VisEQ aligns target-conditioned attention between clean and occluded views during training.}
\end{figure*}

We extend CountGD~\cite{amini2024countgd} to amodal counting through two components. A Feature Reconstruction Module (FRM) reconstructs occlusion-aware feature representations at masked spatial locations where direct visual evidence may be missing. A Visual Equivalence (VisEQ) objective complements this by encouraging, through teacher--student supervision, consistent target-conditioned attention between occluded and unoccluded views during training. At inference, CountOCC uses the input image, target specification, and a predicted occlusion mask. The \textbf{CountOCC} architecture is illustrated in \cref{fig:methodology_overview}.


\subsection{Problem Formulation}

Given an input image $\mathbf{X}_I \in \mathbb{R}^{H \times W \times 3}$ and an occlusion mask $\mathbf{M}_o \in \{0,1\}^{H \times W}$, where $\mathbf{M}_o(i,j)=1$ denotes a spatial location where visual evidence may be missing due to occlusion, our goal is to estimate the amodal count of a target category, including visible, partially visible, and fully occluded instances. The target is specified by visual exemplars $\mathcal{B}=\{\mathbf{b}_1,\ldots,\mathbf{b}_N\}$ and a text description $\mathbf{t}$. During training, $\mathbf{M}_o$ is known from the occluder construction and is used to localize feature reconstruction. At inference, all evaluations use a predicted occlusion mask $\hat{\mathbf{M}}_o$ obtained by the procedure in \cref{sec:inference_mask}, and the model predicts $\hat{y}=f(\mathbf{X}_I,\mathcal{B},\mathbf{t},\hat{\mathbf{M}}_o)$.


\subsection{Feature Reconstruction Module}

Under spatial occlusion, the backbone observes only the visible surface at masked locations and cannot directly encode the missing amodal evidence behind it. We address this limitation with a Feature Reconstruction Module (FRM) that reconstructs occlusion-aware feature representations at masked spatial locations. \cref{fig:feature-reconstruction-module} illustrates the FRM architecture.


\textbf{Visible-Occluded Feature Separation.} Given multi-scale backbone features $\{\mathbf{Z}^{(\ell)}\}_{\ell=1}^{L}$ extracted from the Swin Transformer \cite{liu2021swin} across $L=3$ pyramid levels with channel dimensions $C_{\ell} \in \{256, 512, 1024\}$, we instantiate level-specific feature reconstructors to reconstruct masked representations at each scale. At each level $\ell$, we decompose spatial features into visible and occluded regions according to the occlusion mask $\mathbf{M}_o^{(\ell)}$. Visible tokens $\mathbf{Z}_{\text{vis}}^{(\ell)}$ are directly sampled and flattened from unoccluded regions in the backbone features, while masked positions are represented by learnable query tokens $\mathbf{Q}_0^{(\ell)}$ initialized from a level-specific trainable mask embedding:
\begin{align}
    \mathbf{Z}_{\text{vis}}^{(\ell)} &= flatten(\mathbf{Z}^{(\ell)}[\neg \mathbf{M}_o^{(\ell)}]) \in \mathbb{R}^{B \times N_v \times C_{\ell}} \\
    \mathbf{Q}_0^{(\ell)} &= \text{Replicate}(\boldsymbol{\mu}_{\text{mask}}^{(\ell)}, N_o) \in \mathbb{R}^{B \times N_o \times C_{\ell}}
\end{align}
where $\mathbf{Z}^{(\ell)} \in \mathbb{R}^{B \times C_{\ell} \times H_{\ell} \times W_{\ell}}$ denotes the backbone features at level $\ell$, $\mathbf{M}_o^{(\ell)} \in \{0,1\}^{B \times H_{\ell} \times W_{\ell}}$ represents the downsampled occlusion mask, $N_v$ and $N_o$ indicate the number of visible and occluded tokens respectively, and $\boldsymbol{\mu}_{\text{mask}}^{(\ell)} \in \mathbb{R}^{C_{\ell}}$ is the trainable mask embedding vector specific to pyramid level $\ell$.


\textbf{Spatial-Semantic Attention Fusion.} The occluded features are reconstructed through a sequence of attention operations that integrate spatial context from visible regions and semantic guidance from fused text-visual exemplar embeddings. At each layer, we explicitly inject positional encodings to maintain spatial correspondence across the feature hierarchy. The occluded queries first undergo self-attention \cite{vaswani2017attention} to model interdependencies among masked positions, enabling coherent reconstruction across occluded regions. These refined queries then attend to visible tokens via cross-attention \cite{vaswani2017attention}, propagating contextual information from unoccluded areas, where $\mathbf{\Psi}_{sa}$ and $\mathbf{\Psi}_{ca}$ denote self-attention and cross-attention operations respectively:
\begin{align}
    \mathbf{Q}_{\text{vis}}^{(\ell)} = \mathbf{\Psi}_{ca}(\mathbf{\Psi}_{sa}(\mathbf{Q}_0^{(\ell)}), \mathbf{Z}_{\text{vis}}^{(\ell)}) + \mathbf{\Psi}_{sa}(\mathbf{Q}_0^{(\ell)})
\end{align}

To preserve prompt consistency and reduce reconstruction drift, we further modulate these spatially informed queries through cross-attention over the fused visual-exemplar and text embeddings $\mathbf{Z}_{v,t}$:
\begin{align}
    \mathbf{Z}_{\text{cond}}^{(\ell)} = \mathbf{\Psi}_{ca}(\mathbf{Q}_{\text{vis}}^{(\ell)}, \mathbf{Z}_{v,t}) + \mathbf{Q}_{\text{vis}}^{(\ell)}  
\end{align}

The resulting semantically conditioned queries undergo non-linear transformation via a two-layer MLP, $\mathbf{\Phi}_{mlp}$, with residual skip connection, yielding the reconstructed occluded features:
\begin{align}
    \hat{\mathbf{Z}}_{\text{occ}}^{(\ell)} = \mathbf{\Phi}_{mlp}(\mathbf{Z}_{\text{cond}}^{(\ell)}) + \mathbf{Z}_{\text{cond}}^{(\ell)}
\end{align}


\textbf{Reconstructed Feature Integration.} We reassemble the multi-scale feature maps by preserving visible tokens and inserting reconstructed tokens at masked locations:
\begin{equation}
\tilde{\mathbf{Z}}^{(\ell)} = \begin{cases} 
\mathbf{Z}^{(\ell)} & \text{if } \mathbf{M}_o^{(\ell)} = 0 \\
\hat{\mathbf{Z}}_{\text{occ}}^{(\ell)} & \text{if } \mathbf{M}_o^{(\ell)} = 1
\end{cases}
\end{equation}

The resulting feature pyramids $\{\tilde{\mathbf{Z}}^{(\ell)}\}_{\ell=1}^{L}$ preserve features in visible regions and replace masked locations with reconstructed representations, enabling the cross-modality decoder to identify and enumerate candidate instances through similarity scoring.


\subsection{Reconstruction Loss}

To supervise the FRM, we employ a teacher--student distillation framework~\cite{hinton2015distilling} that anchors reconstructed features to clean representations from an unoccluded teacher network. This encourages reconstructed features to remain aligned with the clean-image feature manifold and discourages drift toward representations inconsistent with the corresponding unoccluded view.

\textbf{Teacher--Student Distillation Setup.} Given an input image $\mathbf{X}_I$ and its synthetically occluded view $\mathbf{X}_{\text{occ}}$ generated according to the occlusion mask $\mathbf{M}_o$, we extract two sets of features under a teacher--student setup during training.
The frozen teacher backbone $f_{\theta_T}$ processes the original unoccluded image to provide reconstruction targets, while the student backbone $f_{\theta_S}$ handles the occluded input:
\begin{align}
    \hat{\mathbf{Z}}_T^{(\ell)} &= f_{\theta_T}^{(\ell)}(\mathbf{X}_I) \odot \mathbf{M}_o^{(\ell)} \label{eq:teacher}\\
    \hat{\mathbf{Z}}_S^{(\ell)} &= \mathcal{R}_{\theta}\big(f_{\theta_S}^{(\ell)}(\mathbf{X}_{\text{occ}}), \mathbf{Z}_{v,t}, \mathbf{M}_o^{(\ell)}\big) \odot \mathbf{M}_o^{(\ell)} \label{eq:student}
\end{align}

where $\hat{\mathbf{Z}}_T^{(\ell)} \in \mathbb{R}^{B \times C_{\ell} \times H_{\ell} \times W_{\ell}}$ denotes the clean teacher features restricted to masked positions, $\hat{\mathbf{Z}}_S^{(\ell)}$ denotes the corresponding student reconstruction, $\mathcal{R}_{\theta}$ is the FRM, and $\mathbf{Z}_{v,t}$ provides guidance from fused visual exemplars and textual prompts. For the reconstruction loss, the teacher path is frozen and gradients are applied only to $\mathcal{R}_{\theta}$, encouraging reconstructed features to align with the clean-image feature manifold.

\textbf{Multi-Term Loss Formulation.} We supervise reconstruction through a composite objective that enforces both geometric accuracy and semantic consistency. Let $\mathcal{O}^{(\ell)} = \{i : \mathbf{M}_o^{(\ell)}[i] = 1\}$ denote the set of occluded spatial positions at level $\ell$. Our loss combines $\ell_2$ distance for magnitude consistency, cosine similarity for angular alignment, and a Charbonnier penalty for robust residual regularization:
\begin{equation} \begin{aligned} \mathcal{L}_{\text{rec}} = \sum_{\ell=1}^{L} \sum_{i \in \mathcal{O}^{(\ell)}} \Bigg[ &\lambda_{\text{charb}} \sqrt{\|\boldsymbol{\Delta}^{(\ell)}_{i}\|_2^2 + \epsilon^2} + \lambda_{\ell_2}\|\boldsymbol{\Delta}^{(\ell)}_{i}\|_2^2 \\ &+ \lambda_{\cos}\left( 1 - \frac{\langle \hat{\mathbf{Z}}_{S,i}^{(\ell)}, \hat{\mathbf{Z}}_{T,i}^{(\ell)} \rangle} {\|\hat{\mathbf{Z}}_{S,i}^{(\ell)}\|_2 \, \|\hat{\mathbf{Z}}_{T,i}^{(\ell)}\|_2 + \epsilon} \right) \Bigg], \end{aligned} \label{eq:reconstruction_loss} \end{equation}

where $\boldsymbol{\Delta}^{(\ell)}_{i}=\hat{\mathbf{Z}}_{S,i}^{(\ell)}-\hat{\mathbf{Z}}_{T,i}^{(\ell)}$ is the reconstruction residual at masked position $i$. This multi-scale supervision enables the reconstruction module to produce features that are metrically accurate and semantically coherent across diverse occlusion patterns.

\subsection{Visual Equivalence}

While FRM recovers occluded features in embedding space, we introduce complementary supervision at the attention level. Our key insight is that the model should attend to consistent target-supporting regions across occluded and unoccluded views. Motivated by SelfEQ~\cite{he2024improved}, we enforce consistency between occluded and unoccluded views through teacher--student supervision. During training, the teacher processes the original image while the student processes the occluded view with reconstructed features, and we align their gradient-based attention maps to encourage consistent localization behavior despite missing visual evidence. \cref{fig:visual-eq} illustrates our proposed VisEQ module.


\textbf{Prompt-Conditioned GradCAM~\cite{selvaraju2017grad}.} We extract multi-level gradient-based attention maps from both teacher processing clean images $\mathbf{X}_I$ and student processing occluded images $\mathbf{X}_{\text{occ}}$. Given final cross-modality decoder similarity scores $\mathbf{Y}\in\mathbb{R}^{B\times Q\times P}$, where $Q$ denotes object queries and $P$ denotes fused prompt tokens, we compute a prompt-conditioned matching score by averaging the top $K$ query scores:
\begin{equation}
\label{eq:matching_score}
s =
\frac{1}{K}
\sum_{i=1}^{K}
\left[
\operatorname{TopK}
\left(
\left\{
\max_{p}\mathbf{Y}_{q,p}
\right\}_{q=1}^{Q},
K
\right)
\right]_{i}
\end{equation}

For each pyramid level $\ell$ with projected features $\mathbf{Z}^{(\ell)} \in \mathbb{R}^{B \times C_\ell \times h_\ell \times w_\ell}$, we compute per-channel importance weights by globally average pooling the gradients of the matching score with respect to each feature channel:
\begin{equation}
\label{eq:gradcam_weights}
\alpha^{(\ell)}_{c} = \frac{1}{h_\ell w_\ell}\sum_{x,y}\frac{\partial s}{\partial \mathbf{Z}^{(\ell)}_{c,x,y}}
\end{equation}

These importance weights are used to compute a weighted combination of feature channels at each level, followed by ReLU activation to retain only positive contributions:
\begin{equation}
\label{eq:level_attention}
\mathbf{\Omega}^{(\ell)} = \text{ReLU}\left(\sum_{c}\alpha^{(\ell)}_{c} \mathbf{Z}^{(\ell)}_{c}\right) \in \mathbb{R}^{h_\ell \times w_\ell}
\end{equation}

Finally, we upsample all pyramid-level attention maps to input resolution and aggregate them using weights $\beta^{(\ell)}$ proportional to gradient energy at each level, producing the final gradient-based attention map $\mathbf{G}$:
\begin{equation}
\label{eq:multilevel_aggregation}
\mathbf{G} = \sum_{\ell=1}^{L} \beta^{(\ell)} \cdot \mathbf{\Omega}^{(\ell)}, \quad \beta^{(\ell)} = \frac{\exp(\sum\left|\frac{\partial s}{\partial \mathbf{Z}^{(\ell)}}\right|)}{\sum_{j=1}^{L}\exp(\sum\left|\frac{\partial s}{\partial \mathbf{Z}^{(j)}}\right|)}
\end{equation}


\subsection{VisEQ Loss}

Our Visual Equivalence supervision operates on attention maps extracted from parallel teacher--student processing paths. For a given input pair $\langle \mathbf{X}_I, \mathbf{X}_{\text{occ}} \rangle$, we extract attention maps $\mathbf{G}_T$ from the teacher network and $\mathbf{G}_S$ from the student network. The maps are conditioned on the fused visual-exemplar and text prompts, yielding target-conditioned attention for the specified category.

\noindent\textbf{Attention Similarity Loss.} We encourage spatial similarity between teacher and student attention maps through a combination of pixel-wise $\ell_2$ distance and cosine similarity. This primary alignment term ensures that the student approximates teacher attention despite processing degraded input:
\begin{equation} \mathcal{L}_{\text{sim}} = \lambda_{\ell_2}\|\mathbf{G}_{T}-\mathbf{G}_{S}\|_2^2 + \lambda_{\cos}\left( 1 - \frac{\langle \mathbf{G}_{T}, \mathbf{G}_{S} \rangle} {\|\mathbf{G}_{T}\|_2 \, \|\mathbf{G}_{S}\|_2 + \epsilon} \right). \label{eq:attention_similarity_loss} \end{equation}
where $\lambda_{\ell_2}$ and $\lambda_{\cos}$ control the relative strength of the pixel-wise and angular alignment terms, respectively.

\noindent\textbf{Region of Interest Consistency.} To prevent trivial solutions where both maps predict uniformly low values, we define a Region of Interest (RoI) mask that identifies spatial locations where at least one network exhibits confident predictions above a threshold $\tau$:
\begin{equation}
\mathbf{M}_{\text{RoI}} = \begin{cases} 1, & \max(\mathbf{G}_{T},\mathbf{G}_{S}) \geq \tau \\ 0, & \text{otherwise} \end{cases}
\end{equation}

Within the RoI, we compute masked attention maps $\mathbf{R}_T = \mathbf{G}_T \odot \mathbf{M}_{\text{RoI}}$ and $\mathbf{R}_S = \mathbf{G}_S \odot \mathbf{M}_{\text{RoI}}$ for teacher and student networks respectively. For each masked map, we compute the mean $\mu_{\text{RoI}}$ and standard deviation $\sigma_{\text{RoI}}$ to capture activation magnitude and consistency. We present the following formulation for the teacher network, with analogous computation for the student network:
\begin{equation}
\mu_{\text{RoI}}^T = \frac{\sum_{i,j} \mathbf{R}_{T,i,j}} {\sum_{i,j} \mathbf{M}_{\text{RoI},i,j}+\epsilon}
\end{equation}
\begin{equation}
\quad \sigma_{\text{RoI}}^T = \sqrt{ \frac{\sum_{i,j} \mathbf{M}_{\text{RoI},i,j} (\mathbf{R}_{T,i,j}-\mu_{\text{RoI}}^T)^2} {\sum_{i,j} \mathbf{M}_{\text{RoI},i,j}+\epsilon}}
\end{equation}

Using the student statistics, we define our consistency loss to encourage high mean activation and low variance within the RoI, where the variance term penalizes inconsistent predictions while the maximum term ensures sufficient activation magnitude, preventing collapse to trivial solutions:

\begin{align}
\mathcal{L}_{\text{cst}}
= \mathbb{E}\Big[
\sigma_{\text{RoI}}^S
+ \max\left(0, \tfrac{\tau}{2} - \mu_{\text{RoI}}^S\right)
\Big]
\end{align}

\subsection{Total Loss}
\label{sec:total_loss}
During training, CountOCC is optimized with the following term:
\begin{equation}
\mathcal{L}_{\text{total}}
=
\mathcal{L}_{\text{count}}
+
\lambda_{\text{rec}}\mathcal{L}_{\text{rec}}
+
\lambda_{\text{sim}}\mathcal{L}_{\text{sim}}
+
\lambda_{\text{cst}}\mathcal{L}_{\text{cst}}
\label{eq:total_loss}
\end{equation}
Here, $\mathcal{L}_{\text{count}}$ is the standard CountGD counting loss, while $\mathcal{L}_{\text{rec}}$, $\mathcal{L}_{\text{sim}}$, and $\mathcal{L}_{\text{cst}}$ supervise feature reconstruction, attention alignment, and RoI-level attention consistency, respectively.


\subsection{Inference-Time Occlusion Mask Construction}
\label{sec:inference_mask}
During inference, given an input image $\mathbf{X}_I$ and target text prompt $\mathbf{t}$, we estimate a predicted occlusion mask $\hat{\mathbf{M}}_o$ using a fixed zero-shot pipeline. We query GroundingDINO~\cite{liu2024grounding} with $\mathbf{t}$ together with generic foreground prompts, e.g., ``foreground object,'' ``object in front,'' and ``occluder,'' to obtain candidate occluding regions. Each candidate box is passed to SAM2~\cite{ravi2025sam} to produce a pixel-level mask. Candidate masks are ranked using detector confidence and monocular depth contrast from Depth Anything V2~\cite{yang2024depth}, filtered by thresholding and non-maximum suppression, and merged to form $\hat{\mathbf{M}}_o$. The predicted mask $\hat{\mathbf{M}}_o$ is used to localize feature reconstruction in CountOCC at inference.

\section{Experiments}
\label{sec:experiments}
We evaluate on four benchmarks covering both controlled and natural occlusion. FSC-147-OCC and CARPK-OCC provide controlled occlusion settings. During training on FSC-147, rectangular masks of $128$-$256$ pixels occlude approximately 15-50\% of target instances, while the fixed validation/test benchmarks occlude approximately 25-35\% of instances with a maximum mask side length of 256 pixels. CARPK-OCC evaluates cross-dataset transfer on overhead parking-lot imagery, whereas KINS~\cite{qi2019amodal} captures naturally occurring inter-object occlusion in real street scenes. We also evaluate on CAPTURe-Real~\cite{pothiraj2025capture}, which contains manually placed occluders in structured repeated-object scenes. Full dataset and construction details are provided in Supplementary Secs.~2 and~5.

We evaluate existing open-world counting methods using their released checkpoints and standard inference protocols. To account for the effect of occlusion-aware training, we additionally include CountGD (Trained), where the original CountGD architecture is trained on the same training split with the same occlusion augmentation used for CountOCC. CountOCC uses predicted occlusion masks $\hat{\mathbf{M}}_o$ generated by the zero-shot pipeline in \cref{sec:inference_mask}. This protocol provides a controlled comparison against both standard open-world counters and an occlusion-trained CountGD baseline under the same evaluation setting.

\textbf{FSC-147-OCC.} 
CountOCC achieves the best overall performance on FSC-147-OCC. It consistently outperforms both text-based methods (CLIP-Count and CounTX) and visual exemplar-based methods (CounTR and LOCA), reducing test MAE by 31.48–51.92\% and RMSE by 48.96–64.84\%. More importantly, occlusion-aware training alone improves CountGD MAE by 7.33\%/7.14\% on validation/test and validation RMSE by 11.20\%, while test RMSE increases by 3.29\%. Relative to this controlled CountGD (Trained) baseline, CountOCC further reduces MAE by 20.86\%/14.19\% and RMSE by 26.65\%/54.63\% on validation/test, respectively. These results show that the gains of CountOCC extend substantially beyond those obtained from exposure to occluded training data alone.

Supplementary Sec.~12 further provides a visible/occluded breakdown on FSC-147-OCC by reporting MAE/RMSE separately for visible and occluded instances within the same image, enabling a finer-grained analysis of performance under occlusion. Supplementary Sec.~13 additionally reports results on the original unoccluded FSC-147 benchmark, showing that CountOCC remains competitive on clean scenes.
\begin{table}[tb]
\centering
\small
\caption{Amodal counting performance on the FSC-147-OCC.}
\setlength{\tabcolsep}{1.5pt}
\begin{tabular}{lcccc}
\toprule
\textbf{Method} & \multicolumn{2}{c}{\textbf{Validation}} & \multicolumn{2}{c}{\textbf{Test}} \\
\cmidrule(lr){2-3} \cmidrule(lr){4-5}
& \textbf{MAE} ↓ & \textbf{RMSE} ↓ & \textbf{MAE} ↓ & \textbf{RMSE} ↓ \\
\midrule
CLIP-Count~\cite{jiang2023clip}  & 26.31&80.45 & 23.90&108.57 \\
CounTX~\cite{AminiNaieni23}  & 24.81&75.58 & 23.04&113.83 \\
\midrule
CounTR~\cite{liu2022countr} & 23.14&66.78 & 22.25&104.75 \\
LOCA~\cite{Dukic_2023_ICCV} & 17.13&44.25 & 16.77&78.41 \\
\midrule
\textsc{CountGD}~\cite{amini2024countgd} & 15.83&54.38 & 14.42&85.40 \\
\textsc{CountGD}~\cite{amini2024countgd} (Trained) & 14.67 &48.29 &13.39 &88.21 \\
\textsc{CountOCC} & \textbf{11.61}&\textbf{35.42} & \textbf{11.49}&\textbf{40.02} \\
\bottomrule
\end{tabular}
\label{tab:fsc147_occlusion_results}
\end{table}

\textbf{CARPK-OCC.}
To assess cross-dataset generalization, we evaluate on CARPK-OCC in a zero-shot setting. CountOCC achieves the best overall performance, reducing MAE/RMSE by 61.29--72.06\%/55.58-67.02\% over the text-based CounTX~\cite{AminiNaieni23} and CLIP-Count~\cite{jiang2023clip}, and by 67.51-77.88\%/59.38-72.14\% over the visual exemplar-based CounTR~\cite{liu2022countr} and LOCA~\cite{Dukic_2023_ICCV}. Compared with CountGD~\cite{amini2024countgd}, occlusion-aware training reduces MAE/RMSE by 32.22\%/23.51\%. CountOCC further improves over this controlled CountGD (Trained) baseline by 22.58\%/20.65\% in MAE/RMSE, demonstrating that the proposed design generalizes beyond the occlusion augmentation used during training.
\begin{table}[tb]
\centering
\small
\caption{Amodal counting performance on the CARPK-OCC test set. $^\dagger$VE: Visual Exemplars.}
\setlength{\tabcolsep}{1pt}
\begin{tabular}{lccc}
\toprule
\textbf{Method} &
\textbf{Input} &
\textbf{MAE} $\downarrow$ &
\textbf{RMSE} $\downarrow$ \\
\midrule
CLIP-Count~\cite{jiang2023clip} & Text & 17.43&20.74 \\
CounTX~\cite{AminiNaieni23} & Text & 12.58&15.40 \\
\midrule
CounTR~\cite{liu2022countr} & VE & 14.99&16.84 \\
LOCA~\cite{Dukic_2023_ICCV} & VE & 22.02&24.55 \\
\midrule
\textsc{CountGD}~\cite{amini2024countgd} & VE \& Text & 9.28&11.27 \\
\textsc{CountGD~\cite{amini2024countgd}} (Trained) & VE \& Text & 6.29 & 8.62 \\
\textsc{CountOCC} & VE \& Text & \textbf{4.87}&\textbf{6.84} \\
\bottomrule
\end{tabular}
\label{tab:carpk_occlusion_results}
\end{table}


\begin{figure*}[!t]
  \centering
  \includegraphics[width=0.90\linewidth,,height=0.28\textheight]{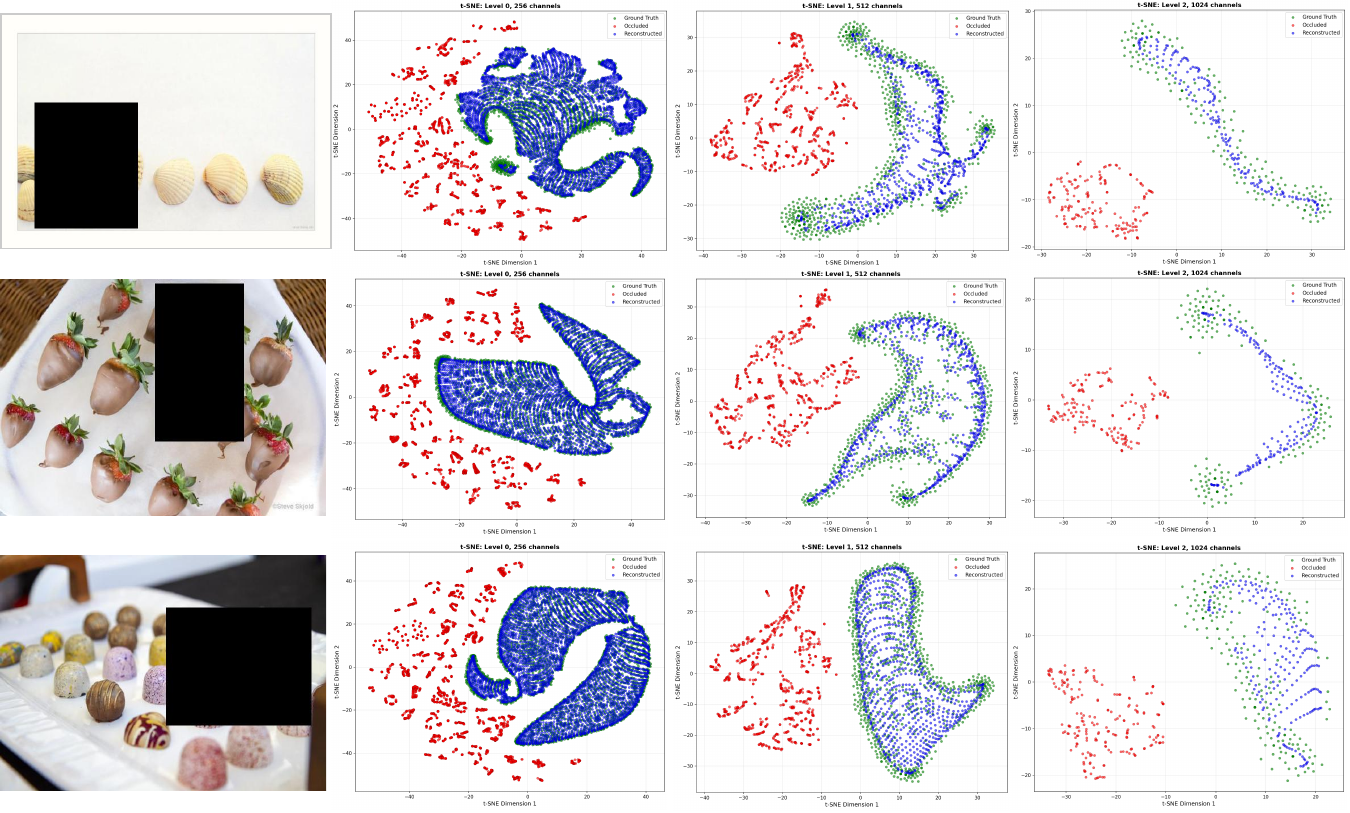}
  \caption{\textbf{Visualization of reconstructed features.}
  Left column shows occluded inputs; remaining columns show t-SNE embeddings
  at three pyramid levels. Occluded, clean ground-truth, and reconstructed
  features are shown in \textcolor{red}{red}, \textcolor{bottlegreen}{green},
  and \textcolor{blue}{blue}, respectively.}
  \label{fig:feature_Viz}
\end{figure*}

\textbf{CAPTURe-Real.}
We further evaluate on CAPTURe-Real~\cite{pothiraj2025capture} in a zero-shot setting. As shown in \cref{tab:capture_results}, CountOCC achieves the best MAE and sMAPE among all evaluated methods, reducing them by 28.52\% and 60.08\% over CountGD, respectively. Relative to CountGD trained on the same occlusion-augmented data, CountOCC further lowers MAE by 5.89\% and sMAPE by 17.41\%, while retaining a similar RMSE (41.38 vs.\ 41.44).
\begin{table}[tb]
\centering
\small
\caption{Amodal counting performance on the CAPTURe-Real.}
\setlength{\tabcolsep}{2pt}
\begin{tabular}{llccc}
\toprule
\textbf{Method} & \textbf{MAE} ↓ & \textbf{RMSE} ↓& \textbf{sMAPE} ↓ \\
\midrule
CLIP-Count~\cite{jiang2023clip}&  16.25&45.78 &26.45  \\
CounTX~\cite{AminiNaieni23}& 20.13&44.73&58.20 \\
\midrule
CounTR~\cite{liu2022countr} & 17.87&51.18&29.74 \\
LOCA~\cite{Dukic_2023_ICCV} &17.28&41.40&36.29 \\
\midrule
\textsc{CountGD}~\cite{amini2024countgd} & 14.97&41.62& 37.68 \\
\textsc{CountGD~\cite{amini2024countgd}} (Trained) &11.37 &41.44&18.21  \\
\textsc{CountOCC}  & \textbf{10.70}&\textbf{41.38}&\textbf{15.04} \\
\bottomrule
\end{tabular}
\label{tab:capture_results}
\end{table}

\textbf{KINS.}
We evaluate on the KINS test split~\cite{qi2019amodal} in a zero-shot setting. Instance annotations are converted into class-wise counting queries over the target categories present in each image. As shown in \cref{tab:kins_results}, CountOCC achieves the best overall performance, reducing MAE/RMSE by 66.23\%/76.45\% over CountGD and by 25.02\%/28.59\% over the matched CountGD (Trained) baseline. Detailed protocol and per-category results are provided in Supplementary Sec.~10.
\begin{table}[tb]
\centering
\small
\caption{Amodal counting performance on the KINS test set.}
\setlength{\tabcolsep}{2pt}
\begin{tabular}{llccc}
\toprule
\textbf{Method} & \textbf{MAE} ↓ & \textbf{RMSE} ↓  \\
\midrule
CLIP-Count~\cite{jiang2023clip}&  53.82&93.99  \\
CounTX~\cite{AminiNaieni23}& 48.81&81.72 \\
\midrule
CounTR~\cite{liu2022countr} & 34.69&51.42 \\
LOCA~\cite{Dukic_2023_ICCV} & 49.28&66.51  \\
\midrule
\textsc{CountGD}~\cite{amini2024countgd}  &31.77 &95.45 \\
\textsc{CountGD~\cite{amini2024countgd}} (Trained) &14.31 &31.48 \\
\textsc{CountOCC}  & \textbf{10.73}&\textbf{22.48} \\
\bottomrule
\end{tabular}
\label{tab:kins_results}
\end{table}


\textbf{Qualitative results.}
Supplementary Sec.~15 provides additional qualitative results on FSC-147-OCC and CARPK-OCC, illustrating CountOCC’s behavior under diverse synthetic occlusion patterns. For each example, we visualize the original image, the occluded image, and the predicted density map, together with ground-truth and predicted counts for total, visible, and occluded instances. These examples show that CountOCC maintains spatially coherent density responses and accurate total-count estimates even when a substantial fraction of target instances is masked. Supplementary Sec.~16 further compares CountOCC with prior open-world counting methods on FSC-147-OCC and CARPK-OCC, and includes a qualitative comparison with CountGD on KINS. These results highlight that existing counters tend to rely primarily on visible evidence, whereas CountOCC remains robust under both controlled synthetic occlusion and naturally occurring occlusion in real street-scene images.

\section{Ablation Study}

\subsection{Design Variants}

We ablate the key components of CountOCC on FSC-147-OCC. As shown in \cref{tab:ablation_frm}, single-level FRM reduces validation MAE by 16.87\%, while multi-level FRM yields larger validation/test MAE reductions of 28.49\%/17.48\%, highlighting the benefit of hierarchical feature reconstruction. However, FRM alone does not consistently reduce RMSE, suggesting that feature recovery primarily improves average counting accuracy but is insufficient for large-error cases. VisEQ alone reduces test MAE/RMSE by 10.61\%/24.66\% relative to the baseline. Combined with multi-level FRM and VisEQ, the complete model achieves the strongest overall performance, reducing validation/test MAE by 26.66\%/20.32\% and RMSE by 34.87\%/53.14\%. This complementary behavior motivates combining feature reconstruction with attention-level consistency.
\begin{table}[!htbp]
\centering
\small
\caption{Design variant analysis on FSC-147-OCC.}
\label{tab:ablation_frm}
\setlength{\tabcolsep}{3pt}
\begin{tabular}{l c c c c}
\toprule
Experiment & \multicolumn{2}{c}{Validation} & \multicolumn{2}{c}{Test} \\
\cmidrule(lr){2-3} \cmidrule(lr){4-5}
 & MAE & RMSE & MAE & RMSE \\
\midrule
Baseline (No FRM/ No VisEQ)&  15.83&54.38&14.42&85.40  \\
FRM (one level)     & 13.16 & 54.51 & 13.77 & 108.63  \\
FRM (all levels)    &\textbf{11.32}&48.12	&11.90&91.45 \\
VisEQ & 13.49& 43.21&12.89 &64.34 \\
FRM (all levels) + VisEQ    & 11.61&\textbf{35.42} & \textbf{11.49}&\textbf{40.02}  \\
\bottomrule
\end{tabular}
\end{table}



\subsection{Visualization of Reconstructed Features}

To validate the effectiveness of our method, we visualize learned representations across network depths using t-SNE dimensionality reduction. \cref{fig:feature_Viz} shows feature embeddings at three pyramid levels. At Level 0, reconstructed features exhibit near-complete overlap with ground truth, demonstrating that our spatial-semantic attention mechanism successfully recovers class-discriminative representations where fine-grained spatial cues are encoded. Occluded features without reconstruction remain distinctly separated from ground truth across all levels. At higher pyramid levels, reconstructed features maintain meaningful alignment with ground truth, although clustering becomes more distributed as features capture increasingly abstract semantics. This progressive recovery validates our hierarchical design, where Level 0 provides the strongest reconstruction impact. Multi-level operation ensures consistent feature recovery across diverse semantic granularities, directly contributing to robust counting under occlusion.

\section{Conclusion}
\label{sec:conclusion and future work}

We introduced \textbf{CountOCC}, an open-world amodal counting framework for single-image spatial occlusion. CountOCC reconstructs occlusion-aware features at masked locations using hierarchical visible-context and text-visual guidance, while VisEQ enforces target-conditioned attention consistency between occluded and unoccluded views. We evaluate on FSC-147-OCC, CARPK-OCC, CAPTURe-Real, and KINS, covering synthetic and natural occlusion. Across these benchmarks, CountOCC consistently improves counting performance, demonstrating the effectiveness of feature reconstruction and attention-level supervision for robust open-world amodal counting.

\clearpage
{
    \small
    \bibliographystyle{ieeenat_fullname}
    \bibliography{1.main}
}
\clearpage
\setcounter{page}{1}
\maketitlesupplementary

\section{Experimental Setup}
\label{sec:experimental setup}

All experiments, including ablation studies, were carried out on a dedicated Linux workstation running Ubuntu 24.04.3 LTS with kernel version 6.14.0-33-generic. The machine is equipped with an Intel Core i9-14900K CPU (24 cores, 32 threads, up to 6.0~GHz), 62~GB of DDR5 RAM, and an NVIDIA GeForce RTX~5090 GPU with 32~GB of VRAM (driver version 580.65.06, CUDA~13.0). The primary storage device is a 1.8~TB NVMe SSD. All code was implemented in Python~3.9.19 and executed within a Conda environment using PyTorch~2.9.0, torchvision~0.25.0, and Transformers~4.39.1. Model training and evaluation were fully GPU-accelerated.

\section{Dataset Details}
\label{sec:dataset details}

\begin{figure}[!htbp]
  \centering
   \includegraphics[width=\linewidth]{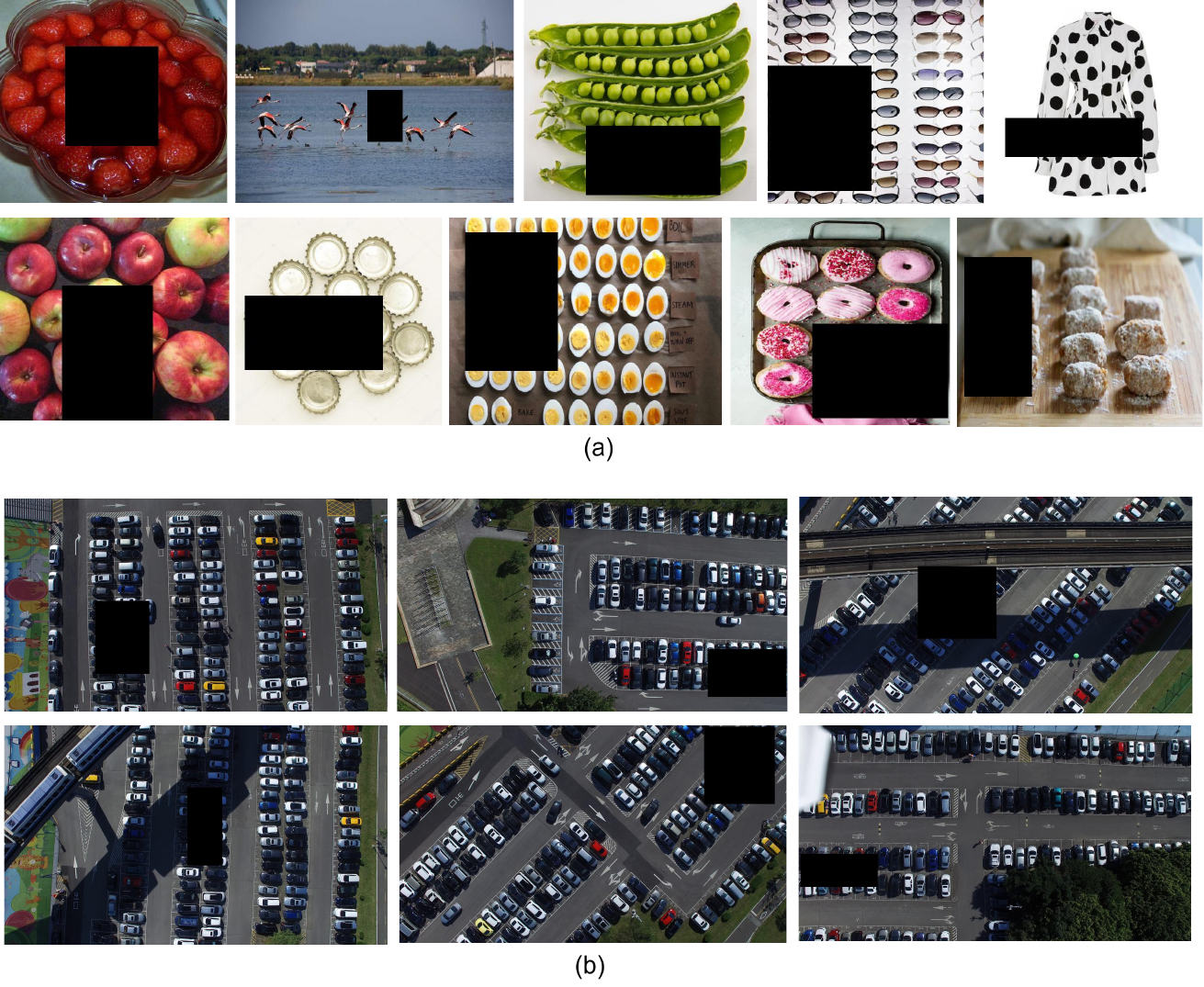}
   \caption{Sample images from (a) FSC-147-OCC and (b) CARPK-OCC benchmarks demonstrating challenging occlusion scenarios for open-world object counting.}
   \label{fig:dataset}
\end{figure}

\begin{figure*}[!htbp]
  \centering
   \includegraphics[width=0.95\linewidth]{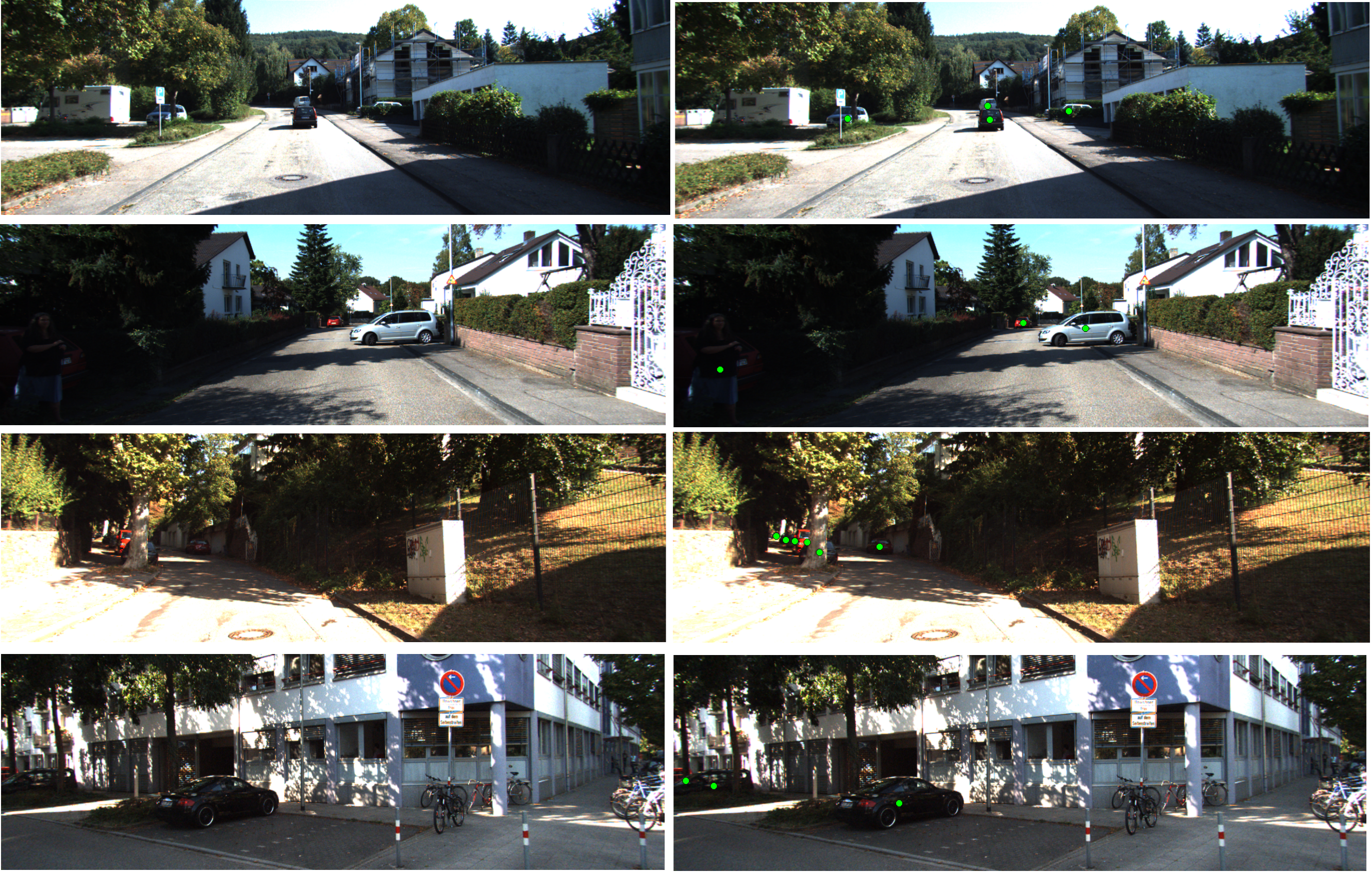}
   \caption{Sample images from the KINS dataset. For each pair, the original image is shown on the left, while the corresponding ground-truth object annotations are shown on the right using \textcolor{bottlegreen}{green} dots. The examples illustrate naturally occurring object occlusions in real-world street scenes.}
   \label{fig:kins_dataset}
\end{figure*}

We evaluate CountOCC on four benchmarks to assess amodal counting under occlusion. For FSC-147~\cite{ranjan2021learning} and CARPK~\cite{hsieh2017drone}, we derive occlusion-augmented variants by applying the training-time and evaluation-time occlusion strategies described in \cref{sec:train_occ} and \cref{sec:eval_occ}, resulting in FSC-147-OCC and CARPK-OCC. These procedures systematically overlay structured rectangular occluders on annotated objects while preserving the original counting annotations, thereby inducing controlled patterns of partial and full occlusion for evaluation. In addition, we use the CAPTURe-Real dataset~\cite{pothiraj2025capture}, which provides structured real-world occlusion settings. We also evaluate on the KINS dataset \cite{qi2019amodal}, which contains naturally occurring occlusions in real street-scene images. Taken together, FSC-147-OCC, CARPK-OCC, CAPTURe-Real, and KINS enable a comprehensive assessment of CountOCC across both synthetically occluded and naturally occluded settings. Representative examples from FSC-147-OCC and CARPK-OCC are shown in \cref{fig:dataset}, while examples from the KINS dataset are shown in \cref{fig:kins_dataset}.

\textbf{FSC-147-OCC.}
FSC-147-OCC is an occlusion-augmented extension of the FSC-147~\cite{ranjan2021learning} dataset that we use for evaluation. During training, occlusion is applied on-the-fly to images from the original FSC-147 training split using the object-aware strategy described in \cref{sec:train_occ}, while preserving all counting annotations. For validation and testing, we construct occlusion-augmented evaluation sets by applying the benchmark occlusion procedure of \cref{sec:eval_occ} to the FSC-147 validation and test splits, yielding FSC-147-OCC-val and FSC-147-OCC-test, respectively.

The underlying FSC-147 dataset is a large-scale open-world counting benchmark comprising 6{,}135 images across 147 object categories, with disjoint class splits for training (89 classes), validation (29), and testing (29). Each image is annotated with object instances and is associated with at least three visual exemplars. For text-based prompts, we follow the FSC-147-D protocol: starting from the original caption, we extract the base noun by removing determiners and modifiers and singularizing the resulting class label; for example, “the donuts in the donut tray” is mapped to the prompt “donut.” To maintain strict comparability with prior work, we also adopt the dataset corrections introduced in CountGD~\cite{amini2024countgd}. In particular, for image \texttt{7171.jpg}, which contains misaligned exemplars, we discard the visual exemplars and retain only the corrected text prompt (“candle”), and for image \texttt{7611.jpg}, where the caption “lego” ambiguously refers to multiple parts, we replace it with the more specific phrase “yellow lego stud” to reflect the intended counting target. In the released COCO-style annotations for FSC-147-OCC, we keep all original instance annotations and additionally record the coordinates of each synthetic occlusion mask, enabling an explicit separation of visible and occluded regions during training, validation, and testing.

\textbf{CARPK-OCC.}
CARPK-OCC is an occlusion-augmented variant of the CARPK dataset, obtained by applying the occlusion strategy from \cref{sec:eval_occ} to the original parking lot images while preserving all counting annotations. The underlying CARPK dataset consists of overhead drone imagery of parking lots with densely arranged vehicles and per-instance bounding-box annotations, comprising 989 training images and 459 test images. We retain the original test split and, following CountGD \cite{amini2024countgd}, use two annotated bounding boxes per image as visual exemplars. For text-based prompts, we adopt the canonical label ``car'' without modification. In the released COCO-style annotations for CARPK-OCC, we keep all original instance annotations and additionally record the coordinates of each synthetic occlusion mask.


\textbf{CAPTURe-Real \cite{pothiraj2025capture}.}
In our experiments, we use the CAPTURe-Real subset of the CAPTURe benchmark, which is specifically designed to assess counting under structured real-world occlusions. CAPTURe-Real consists of 924 images adapted from FSC-147, where human annotators manually place occluders that partially cover repeated object patterns while preserving the underlying counting task. Following the released evaluation metadata used in our experiments, we evaluate on 822 images. Each occluded image is paired with its unoccluded counterpart, enabling a direct measurement of performance degradation under controlled visibility loss. 

\textbf{KINS \cite{qi2019amodal}.}
We further evaluate CountOCC on KINS, a large-scale amodal instance segmentation dataset built from KITTI street-scene images. Although KINS is not originally designed for object counting, we use it to assess zero-shot transfer to naturally occurring inter-object occlusion, complementing the controlled synthetic occlusions in FSC-147-OCC and CARPK-OCC. As illustrated in Fig.~\ref{fig:kins_dataset}, KINS contains real driving scenes with frequent overlap and partial visibility among pedestrians, cyclists, cars, vans, trucks, trams, and other vehicles. The dataset contains $14{,}991$ images, with $7{,}474$ training and $7{,}517$ test images, and provides instance-level category labels together with visible and amodal object annotations. For KINS, ground-truth occlusion support is derived from the annotated instances as the region present in the amodal mask but absent from the corresponding visible mask, with these regions unioned for all instances of the queried category.

For evaluation, we use only the $7{,}517$-image KINS test split; no KINS images are used for training or model selection. We convert the instance annotations into class-wise counting queries by querying each target category present in an image and using the number of annotated instances of that category as the ground-truth count. Thus, an image containing multiple target categories contributes multiple independent counting queries, while categories absent from the image are not queried. This produces $18{,}299$ image--category queries across seven categories. For text-based class specification, we use the corresponding category name as the query, while visual-exemplar specification uses a single visible target bounding box as a one-shot exemplar. Detailed per-category results are reported in \cref{sec:detailed_analysis_kins}.

\section{Evaluation Metrics}
\label{sec:evaluation metrics}

We evaluate counting performance using Mean Absolute Error (MAE), Root Mean Squared Error (RMSE), and, for CAPTURe-Real~\cite{pothiraj2025capture}, symmetric Mean Absolute Percentage Error (sMAPE) following its benchmark protocol. MAE measures the average absolute counting error, while RMSE penalizes larger deviations more strongly and is therefore more sensitive to outliers. sMAPE normalizes the absolute error by the magnitude of the predicted and ground-truth counts, enabling scale-aware comparison across images. Lower values indicate better performance for all three metrics. Let $N$ denote the number of evaluation samples, $\hat{y}_i$ the predicted count, and $y_i$ the ground-truth count for image $X_i$. The metrics are defined as:

\begin{equation}
\text{MAE} =
\frac{1}{N}
\sum_{i=1}^{N}
\left|
\hat{y}_i-y_i
\right|
\end{equation}

\begin{equation}
\text{RMSE} =
\sqrt{
\frac{1}{N}
\sum_{i=1}^{N}
\left(
\hat{y}_i-y_i
\right)^2
}
\end{equation}

\begin{equation}
\text{sMAPE} =
\frac{100}{N}
\sum_{i=1}^{N}
\frac{
\left|
\hat{y}_i-y_i
\right|
}{
\left|
\hat{y}_i
\right|
+
\left|
y_i
\right|
}
\end{equation}

\section{Implementation}
\label{sec:implementation}

\subsection{Training.}

For all experiments, we build on the public CountGD implementation~\cite{amini2024countgd} and closely follow its data augmentation and optimization protocol. Each training image is horizontally flipped with probability $0.5$ and then, with probability $0.5$, either (i) resized so that the shorter side is sampled from $\{480, 512, 544, 576, 608, $
$640, 672, 704, 736, 768, 800\}$ while preserving aspect ratio, or (ii) randomly cropped such that the minimum side lies in $[384, 600]$ and subsequently resized using the same scale set. The resulting image is normalized and passed through the model. Following the FSC-147-D setup, all class names in the FSC-147 training split are concatenated into a single caption, with ``.'' separating class labels, and visual exemplar tokens are appended immediately after the text tokens of their corresponding class. Self-attention masks are constructed so that text tokens attend to each other and to their associated exemplars, but not across unrelated classes. We freeze the Swin-B image encoder and BERT text encoder, and train the projection heads, feature enhancer, cross-modality decoder, and our Feature Reconstruction Module (FRM) and Visual Equivalence (VisEQ) components. Training uses AdamW with weight decay $10^{-4}$, an initial learning rate of $1\times10^{-4}$, and a step-decay schedule over 30 epochs with drops at epochs 10 and 20. Hyperparameters such as localization and classification loss weights, as well as the confidence threshold, follow the best configuration reported for CountGD unless otherwise stated.

On top of this, we adopt a two-stage curriculum tailored to amodal counting under occlusion. In both stages, we train exclusively on the original FSC-147 training split; CARPK is reserved purely for cross-dataset evaluation and is never used during training. In the first stage, we enable only the FRM-related losses and apply the object-aware occlusion augmentation of \cref{sec:train_occ} on-the-fly to each training image. The student network receives synthetically occluded images, while the teacher network processes the corresponding original, unoccluded images. FRM is supervised with a weighted combination of $\ell_2$, cosine, and Charbonnier reconstruction terms, applied over the feature pyramid, but restricted to spatial locations marked as occluded by the binary masks. In this stage, model selection is performed on the FSC-147-OCC validation split, obtained by applying the evaluation-time occlusion scheme of \cref{sec:eval_occ} to the FSC-147 validation set.

In the second stage, we initialize from the best FRM checkpoint and jointly train FRM and VisEQ while continuing to use the same training-time occlusion augmentation on FSC-147. For every 5 training iterations, we compute Grad-CAM-style, language-conditioned attention maps for the original (teacher) and occluded (student) views and impose our similarity and self-consistency losses on these maps, in addition to the standard detection loss and FRM reconstruction loss. This two-stage procedure first stabilizes feature reconstruction under occlusion and then refines the alignment between the model’s responses to visible and occluded inputs.

We inherit the Feature Enhancer ($f_{\phi}$), Top-$K$ query selection, and Cross-Modality Decoder ($f_{\psi}$) from CountGD~\cite{amini2024countgd} with the same architecture and default hyperparameters. CountOCC-specific optimization and loss hyperparameters are summarized in Tab.~\ref{tab:hyperparameters}. The final model is selected based on performance on the FSC-147-OCC validation set and is subsequently evaluated on the FSC-147-OCC test split, CARPK-OCC, CAPTURe-Real, and KINS.

\subsection{Inference. }
At inference time, we deploy only the student branch of CountOCC. Each test image is resized so that its shorter side is 800 pixels while preserving aspect ratio, normalized, and processed together with the text query and visual exemplars. Importantly, CountOCC does not use benchmark-provided occlusion masks as model input. Instead, given the input image and prompt, we first estimate an occlusion mask $\hat{M}_{o}$ using our zero-shot occlusion-mask construction pipeline in Sec. \ref{supp:zeroshot-occ-mask}. 


The image is then passed through the frozen Swin-B image encoder to obtain a multi-scale feature pyramid. In the student branch, FRM reconstructs features only inside the predicted occlusion support $\hat{M}_{o}$, while leaving features outside this region unchanged. The completed feature maps are subsequently fed into the multi-modal decoder to produce cross-modality queries. For each query, we compute the maximum similarity over all text and exemplar tokens and retain predictions whose score exceeds a fixed confidence threshold. Our implementation also supports optional adaptive cropping and SAM-based test-time foreground normalization; for CountGD-based comparisons, these options are kept identical across methods. The teacher path of FRM and all VisEQ-related components are used exclusively during training and are fully disabled at test time.

The total count is obtained by enumerating all retained detections above the confidence threshold. For diagnostic analysis, when region-level annotation masks are available, we additionally compute visible and occluded counts by intersecting detection centers with the annotated visible/occluded regions, yielding $y_{\text{vis}}$, $y_{\text{occ}}$, and $y_{\text{total}} = y_{\text{vis}} + y_{\text{occ}}$. These annotation masks are used only for constructing synthetic occluded images during dataset preparation and for reporting visible/occluded breakdowns during evaluation. They are never provided to CountOCC as oracle input at inference time. All quantitative and qualitative results reported in the main paper follow this unified inference protocol.

\subsection{Hyperparameter Settings}
\label{supp:hyperparameter}
Table~\ref{tab:hyperparameters} summarizes the main CountOCC-specific hyperparameters used in our experiments.

\begin{table}[t]
\centering
\caption{Main hyperparameter settings used for CountOCC.}
\label{tab:hyperparameters}
\begin{tabular}{lc}
\toprule
\textbf{Hyperparameter} & \textbf{Value} \\
\midrule
Batch size & 2 \\
Training epochs & 30 \\
Learning rate & $1\times10^{-4}$ \\
Weight decay & $1\times10^{-4}$ \\
FRM decoder layers & 2 \\
FRM attention heads & 8 \\
Reconstruction $\ell_2$ weight $(\lambda_{\ell_2})$ & 0.3 \\
Reconstruction cosine weight $(\lambda_{\mathrm{cos}})$ & 0.3 \\
Reconstruction Charbonnier weight $(\lambda_{\mathrm{charb}})$ & 0.3 \\
Reconstruction loss weight $(\lambda_{\mathrm{rec}})$ & 0.3 \\
VisEQ similarity weight $(\lambda_{\mathrm{sim}})$ & 1.0 \\
VisEQ consistency weight $(\lambda_{\mathrm{cst}})$ & 1.0 \\
Attention threshold $(\tau)$ & 0.7 \\
Top-$K$ queries $(K)$ & 900 \\
Numerical stability constant $(\epsilon)$ & $1\times10^{-6}$ \\
Occlusion probability ($p$) & 0.6 \\
\bottomrule
\end{tabular}
\end{table}

\section{Occlusion Strategy}
\label{sec:occlusiorn strategy}

\subsection{Occlusion Strategy during Training-Time}
\label{sec:train_occ}

During training, we do not rely on the pre-generated occlusion benchmarks directly. Instead, we apply an on-the-fly, object-aware occlusion augmentation to the FSC-147 training split. This procedure leverages ground-truth bounding boxes to synthesize rectangular occluders while explicitly controlling the fraction of objects that are occluded in each image.

Given an input image $X \in \mathbb{R}^{3 \times H \times W}$ from FSC-147 and its target annotation $t$, we first map the normalized ground-truth boxes to pixel coordinates and compute the number of annotated instances $N$. With probability $p$, we apply occlusion; with the remaining probability $1 - p$, instead, we use an all-zero mask and keep the image unchanged. When occlusion is applied, we target a controlled proportion of instances by selecting a number of objects between $N_{\min} = \lceil \alpha_{\min} N \rceil$ and $N_{\max} = \lfloor \alpha_{\max} N \rfloor$, with $\alpha_{\min} = 0.15$ and $\alpha_{\max} = 0.50$. For images containing very few instances ($N < 4$), this range is clamped so that at least one and at most two objects are occluded, thereby avoiding degenerate cases.

To construct an occlusion mask, we iteratively sample candidate rectangular occluders. At each trial, we (i) randomly select a ground-truth bounding box as an anchor, (ii) place the occluder so that its center coincides with the center of this box in image coordinates, and (iii) sample its height and width uniformly between 128 and 256 pixels, further clipped to respect the image boundaries. The resulting rectangle is then constrained to lie fully within the image, and all pixels inside it are marked as occluded in a candidate mask $\tilde{M} \in \{0,1\}^{H \times W}$. Using the projected object centers, we count how many instances fall inside $\tilde{M}$; if this count lies within the desired range $[N_{\min}, N_{\max}]$, we accept the candidate and set $M = \tilde{M}$. If no valid candidate is found within a fixed number of attempts (50 in practice), we instead sample a rectangle with the same size constraints but at a random image location, yielding an occluder that is no longer explicitly anchored to a particular object.

The final binary mask $M$ is stored as the occlusion mask in the training target and is used to construct the student input $\tilde{X}$ by overwriting masked pixels with a black mask. The teacher network always processes the original, unoccluded image $X$, while the student receives it $\tilde{X}$ together with $M$. All FRM losses are computed only at feature locations whose spatial coordinates fall inside the occluded region (i.e., where $M = 1$), encouraging the model to reconstruct missing instance features.

This training-time strategy is (i) \emph{object-aware}, since occluders are anchored on annotated objects and constrained to cover a controlled fraction of them; (ii) \emph{diverse}, as masks are sampled independently per image and iteration with randomized patch sizes and positions; and (iii) \emph{decoupled} from evaluation, since FSC-147-OCC is generated once from the validation and test splits and used only for validation and testing, not for online augmentation.

\subsection{Occlusion Strategy for Evaluation Benchmarks}
\label{sec:eval_occ}

To systematically evaluate amodal counting under controlled occlusion, we construct occlusion-augmented evaluation sets for both FSC-147 and CARPK. For each dataset, which provides object-level bounding-box annotations, we synthesize structured occluders by overlaying black rectangular masks whose centers are aligned with annotated objects, targeting approximately 25--35\% of the instances in each image. This masking strategy provides fine-grained control over how many objects become partially or fully hidden while preserving the original count annotations.

For FSC-147, we apply this procedure separately to the validation and test splits, yielding two occlusion-augmented sets: FSC-147-OCC-val and FSC-147-OCC-test. The former is derived from the original FSC-147 validation images and is used during training only for model selection, whereas FSC-147-OCC-test is held out strictly for final evaluation. For each object, we restrict occluders to have a maximum side length of 256 pixels and select, among all valid candidates, the rectangle whose overlap with the object box most closely matches the target occlusion ratio under the image-size constraints. The resulting occlusion masks are applied directly to the RGB images, and the corresponding metadata is stored in updated COCO-style annotations, which we use to derive visible and occluded subsets during validation and testing.

For CARPK, we follow the same center-based masking scheme on the official test split to construct CARPK-OCC-test, again targeting 25--35\% occlusion of annotated vehicles with a maximum occluder side length of 256 pixels. CARPK-OCC-test is used solely as a cross-dataset evaluation benchmark; the model is never trained on CARPK images. During validation and test-time inference, all quantitative and qualitative results are reported on FSC-147-OCC-val, FSC-147-OCC-test, and CARPK-OCC-test, while training relies only on the clean FSC-147 training split with on-the-fly occlusion augmentation (\cref{sec:train_occ}).

\subsection{Zero-Shot Occlusion Mask Prediction}
\label{supp:zeroshot-occ-mask}

During inference time, we estimate a predicted occlusion support $\hat{M}_{o}$ before running the counting model. This stage is used only to identify image regions where feature reconstruction should be applied by FRM.

Given an image and the counting prompt, we construct a compact set of occlusion-oriented text prompts that describe likely foreground blockers, optionally augmented with the target category name when available. GroundingDINO \cite{liu2024grounding} is then used to localize candidate occluder regions from these prompts. We apply fixed box and text confidence thresholds to remove low-confidence detections before passing the remaining boxes to SAM2 \cite{ravi2025sam} for mask prediction.

Each retained GroundingDINO box is passed to SAM2 as a box prompt. When multiple masks are returned, we select the mask with the highest SAM2-predicted quality score. We then assign each candidate mask a scalar score based on detector confidence, SAM2 mask quality, mask geometry, image contrast, and relative depth contrast from Depth Anything V2. The geometric terms remove masks that are too small or cover an implausibly large fraction of the image, while favoring compact foreground regions. The depth term measures the absolute contrast between the average relative depth inside and outside the mask; therefore, it does not assume a fixed near/far direction for the monocular depth map.

After scoring, candidates below the score threshold are discarded. The remaining masks are sorted by score and filtered using non-maximum suppression with box IoU threshold of $0.5$ and a containment threshold of $0.8$. The final support $\hat{M}_{o}$ is obtained by taking the union of the retained masks, followed by connected-component cleanup. For compatibility with the FRM input format, each connected component is converted to a rectangular binary mask. The predicted masks are precomputed once for each evaluation split and cached before CountOCC inference.

Dataset annotation masks, when available, are used only for constructing synthetic occluded images and for reporting diagnostic visible/occluded-region metrics. They are not provided to CountOCC as oracle occlusion masks during inference.

\begingroup
\small
\setlength{\parindent}{0pt}
\setlength{\parskip}{0pt}

\noindent\rule{\linewidth}{0.4pt}
\textbf{Algorithm 1. Zero-Shot Occlusion Support Estimation}
\vspace{0.2em}

\begin{algorithmic}[1]
\Require Image $I$, counting prompt $q$
\Ensure Predicted occlusion support $\hat{M}_{o}$

\State $\mathcal{P}_{o} \gets$ build occlusion-oriented prompts from generic foreground terms and $q$
\State $\mathcal{B} \gets \textsc{GroundingDINO}(I,\mathcal{P}_{o})$
\State $\mathcal{B} \gets$ filter low-confidence and invalid boxes
\State $\mathcal{C} \gets \emptyset$

\For{each box $b \in \mathcal{B}$}
\State $m \gets$ best SAM2 mask using $b$ as box prompt
\State $s \gets \mathrm{score}(m)$ using confidence, mask quality, geometry, contrast, and depth contrast
\If{$m$ passes score and geometric filters}
\State $\mathcal{C} \gets \mathcal{C} \cup {(m,s)}$
\EndIf
\EndFor

\State $\mathcal{C} \gets$ sort by score and apply NMS
\State $\mathcal{C} \gets$ keep top-ranked occlusion components
\State $\hat{M}_{o} \gets \bigcup{(m,s)\in\mathcal{C}} m$
\State $\hat{M}_{o} \gets$ connected-component cleanup$(\hat{M}_{o})$
\State Convert components of $\hat{M}_{o}$ to FRM-compatible binary masks
\State Cache $\hat{M}_{o}$ for CountOCC inference
\State \Return $\hat{M}_{o}$
\end{algorithmic}

\noindent\rule{\linewidth}{0.4pt}
\endgroup

\subsection{Occlusion-Mask Generation Runtime}
\label{supp:runtime}
The zero-shot mask-generation pipeline requires an average of 637.48\,ms per image. Each predicted mask is generated once per evaluation image and cached before CountOCC inference, avoiding repeated mask-generation overhead during counting.

\section{Effect of Occlusion-Mask Source}
\label{sec:mask_source_ablation}

To assess the effectiveness of the zero-shot occlusion-mask pipeline used at inference time, we compare CountOCC with ground-truth occlusion masks and zero-shot predicted masks while keeping the counting model fixed. For KINS, the ground-truth support is obtained from the difference between the amodal and visible instance masks described in \cref{sec:dataset details}. As shown in \cref{tab:mask_source_ablation}, the predicted masks produce results that are consistently close to those obtained with ground truth masks across FSC-147-OCC, CARPK-OCC, CAPTURe-Real, and KINS. This demonstrates that the zero-shot pipeline provides reliable occlusion support for feature reconstruction and enables CountOCC to maintain strong amodal counting performance at inference time.


\begin{table}[!htbp]
\centering
\small
\setlength{\tabcolsep}{4pt}
\caption{Effect of occlusion-mask source on CountOCC.}
\begin{tabular}{llcc}
\toprule
Dataset / Split & Mask source & MAE & RMSE \\
\midrule
FSC-147-OCC Val  & Ground Truth mask                  & 11.60 & 35.40 \\
                 & Zero-shot predicted mask  & 11.61 & 35.42 \\
\midrule
FSC-147-OCC Test & Ground Truth mask                  & 11.42 & 38.68 \\
                 & Zero-shot predicted mask  & 11.49 & 40.02 \\
\midrule
CARPK-OCC Test   & Ground Truth mask                  & 4.65  & 5.91  \\
                 & Zero-shot predicted mask  & 4.87  & 6.84  \\
\midrule
CAPTURe-Real     & Ground Truth mask                  & 10.66 & 41.31 \\
                 & Zero-shot predicted mask  & 10.70 & 41.38 \\
\midrule
KINS Test             & Ground Truth mask                  & 10.68  & 22.40 \\
                 & Zero-shot predicted mask  & 10.73  & 22.48 \\
\bottomrule
\end{tabular}
\label{tab:mask_source_ablation}
\end{table}


\section{Occlusion-Mask Prediction Quality}
\label{sec:mask_quality}

To quantitatively evaluate the zero-shot occlusion-mask prediction pipeline, we compare the predicted masks with the available annotation masks using intersection-over-union (IoU), precision, recall, and Dice score. As shown in \cref{tab:mask_quality}, the predicted masks achieve consistently high overlap on FSC-147-OCC, CARPK-OCC, and CAPTURe-Real, with IoU ranging from 0.92 to 0.96 and Dice from 0.94 to 0.97. Performance is lower on KINS, where occlusions arise naturally from complex inter-object interactions, but remains robust with an IoU of 0.75 and Dice of 0.86. These results provide direct quantitative validation of the predicted occlusion support used by CountOCC at inference time and complement the mask-source analysis in \cref{sec:mask_source_ablation}.

\begin{table}[!htbp]
\centering
\small
\caption{Quality of the zero-shot predicted occlusion masks. $n$ denotes the number of evaluation samples; for KINS, each image category query is treated as one evaluation sample. Higher values indicate better agreement with the annotation masks.}
\setlength{\tabcolsep}{2.5pt}
\begin{tabular}{lccccc}
\toprule
\textbf{Dataset / Split} &
\textbf{$n$} &
\textbf{IoU} &
\textbf{Precision} &
\textbf{Recall} &
\textbf{Dice} \\
\midrule
FSC-147-OCC Val.  & 1286 & 0.95 & 0.95 & 0.99 & 0.97 \\
FSC-147-OCC Test  & 1190 & 0.96 & 0.95 & 0.99 & 0.97 \\
CARPK-OCC         & 459  & 0.94 & 0.94 & 0.98 & 0.96 \\
CAPTURe-Real      & 822  & 0.92 & 0.92 & 0.96 & 0.94 \\
KINS Test              & 18299   & 0.75 & 0.85 & 0.88 & 0.86 \\
\bottomrule
\end{tabular}
\label{tab:mask_quality}
\end{table}

\section{Sensitivity to Occlusion-Mask Perturbations}
\label{sec:mask_perturbation}

To evaluate sensitivity to errors in the predicted occlusion support, we perturb the predicted masks only at inference time while keeping the trained CountOCC model and all other inference settings fixed. For the \emph{shifted} setting, each connected mask component is translated by 10\% of its bounding-box width or height in each of the four cardinal directions, with results averaged across directions and regions extending beyond the image boundary clipped accordingly. This preserves the mask area while introducing controlled localization errors. For the \emph{eroded} and \emph{dilated} settings, we apply morphological erosion or dilation once using a structuring-element radius equal to 5\% of the shorter side of the corresponding mask bounding box, simulating under- and over-estimation of the occluded support, respectively. As shown in \cref{tab:mask_perturbation}, spatial shifts cause only modest degradation, while erosion and particularly dilation produce larger errors. The stronger effect of dilation indicates that reconstructing unnecessarily broad image regions is more detrimental than moderate localization error or conservative mask underestimation. Overall, performance degrades gradually under these perturbations, showing that CountOCC does not require exact mask boundaries and remains robust to reasonable errors in the predicted occlusion support.

\begin{table*}[!htbp]
\centering
\small
\caption{Sensitivity of CountOCC to perturbations of the predicted occlusion mask. Lower MAE and RMSE are better.}
\setlength{\tabcolsep}{1.4pt}
\begin{tabular}{lcccccccc}
\toprule
& \multicolumn{4}{c}{\textbf{FSC-147-OCC}}
& \multicolumn{2}{c}{\textbf{CARPK-OCC}}
& \multicolumn{2}{c}{\textbf{CAPTURe-Real}} \\
\cmidrule(lr){2-5}
\cmidrule(lr){6-7}
\cmidrule(lr){8-9}

& \multicolumn{2}{c}{\textbf{Validation}}
& \multicolumn{2}{c}{\textbf{Test}}
& \multicolumn{2}{c}{}
& \multicolumn{2}{c}{} \\
\cmidrule(lr){2-3}
\cmidrule(lr){4-5}

\textbf{Mask}
& \textbf{MAE} $\downarrow$
& \textbf{RMSE} $\downarrow$
& \textbf{MAE} $\downarrow$
& \textbf{RMSE} $\downarrow$
& \textbf{MAE} $\downarrow$
& \textbf{RMSE} $\downarrow$
& \textbf{MAE} $\downarrow$
& \textbf{RMSE} $\downarrow$ \\
\midrule

Predicted
& \textbf{11.61} & \textbf{35.42}
& \textbf{11.49} & \textbf{40.02}
& \textbf{4.87}  & \textbf{6.84}
& \textbf{10.70} & \textbf{41.38} \\

Shifted
& 11.94 & 36.48
& 11.88 & 42.10
& 5.08  & 7.12
& 10.91 & 41.76 \\

Eroded
& 12.32 & 38.06
& 12.37 & 45.18
& 5.41  & 7.61
& 11.28 & 42.34 \\

Dilated
& 13.08 & 42.73
& 13.16 & 57.64
& 6.24  & 8.93
& 12.46 & 44.17 \\

\bottomrule
\end{tabular}
\label{tab:mask_perturbation}
\end{table*}

\section{Comparison with inpainting}
\label{sec:inpainting}

To further evaluate whether mask access alone is sufficient for amodal counting, we compare CountOCC with a same-mask RGB inpainting baseline. In this setting, SDXL-inpaint~\cite{podell2024sdxl} + CountGD uses the same occlusion mask $\hat{M}_{o}$ to fill the masked image region with Stable Diffusion XL inpainting, after which the completed image is passed to CountGD using its standard counting pipeline. As shown in \cref{tab:inpainting}, this same-mask inpainting baseline remains substantially below CountOCC on both FSC-147-OCC and CARPK-OCC. These results show that inpainting the masked pixels is not sufficient to recover the counting-relevant evidence needed for amodal enumeration and that the inpainting-based solution cannot match CountOCC’s feature-level reconstruction performance.

\begin{table}[!htbp]
\centering
\footnotesize
\setlength{\tabcolsep}{1.4pt}
\caption{Same-mask RGB-inpainting baseline. Both methods use the same occlusion mask; SDXL-inpaint + CountGD first completes the masked RGB region and then applies CountGD.}
\begin{tabular}{llcccc}
\toprule
Dataset & Method & \multicolumn{2}{c}{Validation} & \multicolumn{2}{c}{Test} \\
\cmidrule(lr){3-4}\cmidrule(lr){5-6}
& & MAE$\downarrow$ & RMSE$\downarrow$ & MAE$\downarrow$ & RMSE$\downarrow$ \\
\midrule
FSC-147-OCC & SDXL-inpaint+CountGD & 14.52 & 52.60 & 13.78 & 88.03 \\
            & CountOCC                & \textbf{11.61} & \textbf{35.42} & \textbf{11.49} & \textbf{40.02} \\
\midrule
CARPK-OCC   & SDXL-inpaint+CountGD & -- & -- & 9.33 & 11.80 \\
            & CountOCC                & -- & -- & \textbf{4.87} & \textbf{6.84} \\
\bottomrule
\end{tabular}
\label{tab:inpainting}
\end{table}

\section{Per-Category Analysis on KINS}
\label{sec:detailed_analysis_kins}

To provide a more detailed evaluation under naturally occurring occlusion, we report per-category results on KINS in \cref{tab:detailed_analysis_kins}. No KINS images are used for training; CountGD, CountGD (Trained), and CountOCC are evaluated using category-name text queries only, whereas CounTR and LOCA use the one-shot visual-exemplar protocol described in \cref{sec:dataset details}. In addition to the original CountGD, we include CountGD (Trained), which is trained on FSC-147 using the same occlusion augmentation as CountOCC, thereby separating the benefit of occlusion-aware training from that of the proposed architecture. Occlusion-aware training substantially improves the aggregate CountGD baseline from 31.77/95.45 to 14.31/31.48 MAE/RMSE, while CountOCC further reduces these errors to 10.73/22.48. CountOCC also achieves lower MAE than both CountGD baselines across all evaluated categories, including pedestrians, cyclists, cars, vans, trucks, trams, and miscellaneous road-scene objects. These results show that the improvement on KINS is not explained by exposure to synthetic occlusion alone and that the proposed occlusion-aware reconstruction transfers effectively to natural inter-object occlusion.

\begin{table*}[!htbp]
\centering
\small
\setlength{\tabcolsep}{3.5pt}
\caption{Per-category amodal counting performance on KINS. MAE/RMSE are reported for total-count prediction.}
\begin{tabular}{lccccccc}
\toprule
\textbf{Category} & \textbf{Queries}
& \multicolumn{2}{c}{\textsc{CountGD}~\cite{amini2024countgd}}
& \multicolumn{2}{c}{\textsc{CountGD}~\cite{amini2024countgd} (Trained)}
& \multicolumn{2}{c}{\textsc{CountOCC}} \\
\cmidrule(lr){3-4}
\cmidrule(lr){5-6}
\cmidrule(lr){7-8}

& &
\textbf{MAE}$\downarrow$ & \textbf{RMSE}$\downarrow$ &
\textbf{MAE}$\downarrow$ & \textbf{RMSE}$\downarrow$ &
\textbf{MAE}$\downarrow$ & \textbf{RMSE}$\downarrow$ \\
\midrule

Car
& 7006
& 5.87 & \textbf{11.37}
& 4.80 & 14.00
& \textbf{4.03} & 16.38 \\

Pedestrian
& 3231
& 76.44 & 170.25
& 24.00 & 42.00
& \textbf{16.14} & \textbf{20.93} \\

Cyclist
& 1446
& 19.40 & 22.86
& 14.50 & 22.50
& \textbf{12.85} & \textbf{21.94} \\

Van
& 3248
& 11.77 & 14.55
& 9.50 & 12.00
& \textbf{8.22} & \textbf{9.78} \\

Truck
& 597
& 13.43 & 15.89
& 11.00 & 13.50
& \textbf{9.61} & \textbf{11.09} \\

Tram
& 490
& 17.61 & 21.54
& 11.50 & 15.50
& \textbf{9.83} & \textbf{12.44} \\

Misc.
& 2281
& 92.29 & 175.64
& 38.00 & 65.00
& \textbf{26.38} & \textbf{45.86} \\

\midrule
\textbf{Micro Avg.}
& 18299
& 31.77 & 95.45
& 14.31 & 31.48
& \textbf{10.73} & \textbf{22.48} \\

\textbf{Macro Avg.}
& --
& 33.83 & 61.73
& 16.19 & 26.36
& \textbf{12.44} & \textbf{19.77} \\

\bottomrule
\end{tabular}
\label{tab:detailed_analysis_kins}
\end{table*}

\section{Performance Across Occlusion Severity}
\label{sec:severity_analysis}

To evaluate robustness under varying degrees of occlusion, we group the evaluation samples into five severity ranges according to the fraction of target evidence occluded and report MAE in \cref{tab:severity_analysis}. CountOCC consistently improves over CountGD (Trained), which is trained with the same occlusion augmentation, across all available severity levels on FSC-147-OCC, CARPK-OCC, and CAPTURe-Real. FSC-147-OCC provides the strongest evidence under severe occlusion: at 60-80\% severity, CountOCC reduces MAE from 27.56 to 18.77 on validation and from 17.52 to 13.60 on test, with improvements also observed in the 80--100\% bin. The highest-severity bins of CARPK-OCC and CAPTURe-Real contain very few samples and are therefore reported for completeness rather than used to draw strong conclusions. These results show that the gains are not limited to mild occlusion or exposure to occluded training samples but persist as increasingly more target evidence becomes unavailable.

\begin{table}[t]
\centering
\footnotesize
\caption{Counting performance across occlusion severity levels. Results are reported in MAE; $n$ denotes the number of evaluation samples in each severity bin.}
\setlength{\tabcolsep}{1.6pt} \begin{tabular}{llrrrr} \toprule \textbf{Dataset / Split} & \textbf{Severity(\%)} & \textbf{$n$} & \textbf{CountGD} & \begin{tabular}[c]{@{}c@{}}
\textbf{CountGD}\\
\textbf{(Trained)}
\end{tabular} &\textbf{CountOCC} \\ \midrule

\multirow{6}{*}{FSC-147 Val}
& 0-20   & 308  & 14.93 & 14.53 & \textbf{12.28} \\
& 20-40  & 651  & 11.03 & 10.73 & \textbf{8.75} \\
& 40-60  & 148  & 8.43  & 10.27 & \textbf{8.20} \\
& 60-80  & 153  & 36.26 & 27.56 & \textbf{18.77} \\
& 80-100 & 26   & 68.38 & 64.14 & \textbf{52.56} \\
& All       & 1286 & 15.83 & 14.67 & \textbf{11.61} \\
\midrule

\multirow{6}{*}{FSC-147 Test}
& 0-20   & 248  & 24.62 & 24.20 & \textbf{22.86} \\
& 20-40  & 580  & 9.81  & 8.45  & \textbf{7.01} \\
& 40-60  & 180  & \textbf{7.09} & 9.66 & 7.28 \\
& 60-80  & 151  & 23.64 & 17.52 & \textbf{13.60} \\
& 80-100 & 31   & \textbf{16.75} & 20.89 & 18.68 \\
& All       & 1190 & 14.42 & 13.39 & \textbf{11.49} \\
\midrule

\multirow{6}{*}{CARPK-OCC}
& 0-20   & 420 & 9.15  & 6.22 & \textbf{4.84} \\
& 20-40  & 38  & 10.65  & 7.01 & \textbf{5.22} \\
& 40-60  & 1   & 12.53 & 8.21 & \textbf{5.78} \\
& 60-80  & -- & -- & -- & -- \\
& 80-100 & --  & --    & --   & -- \\
& All       & 459 & 9.28  & 6.29 & \textbf{4.87} \\
\midrule

\multirow{6}{*}{CAPTURe-Real}
& 0-20   & 177  & 13.32 & 10.25  & \textbf{9.95} \\
& 20-40  & 534  & 14.86 & 11.26 & \textbf{10.63} \\
& 40-60  & 108  & 18.07 & 13.65 & \textbf{12.21} \\
& 60-80  & 3    & 20.21 & 14.95 & \textbf{13.57} \\
& 80-100 & --  & -- & -- & -- \\
& All       & 822 & 14.97 & 11.37 & \textbf{10.70} \\

\bottomrule
\end{tabular}
\label{tab:severity_analysis}
\end{table}

\begin{figure*}[!tb]
  \centering
   \includegraphics[width=0.95\linewidth]{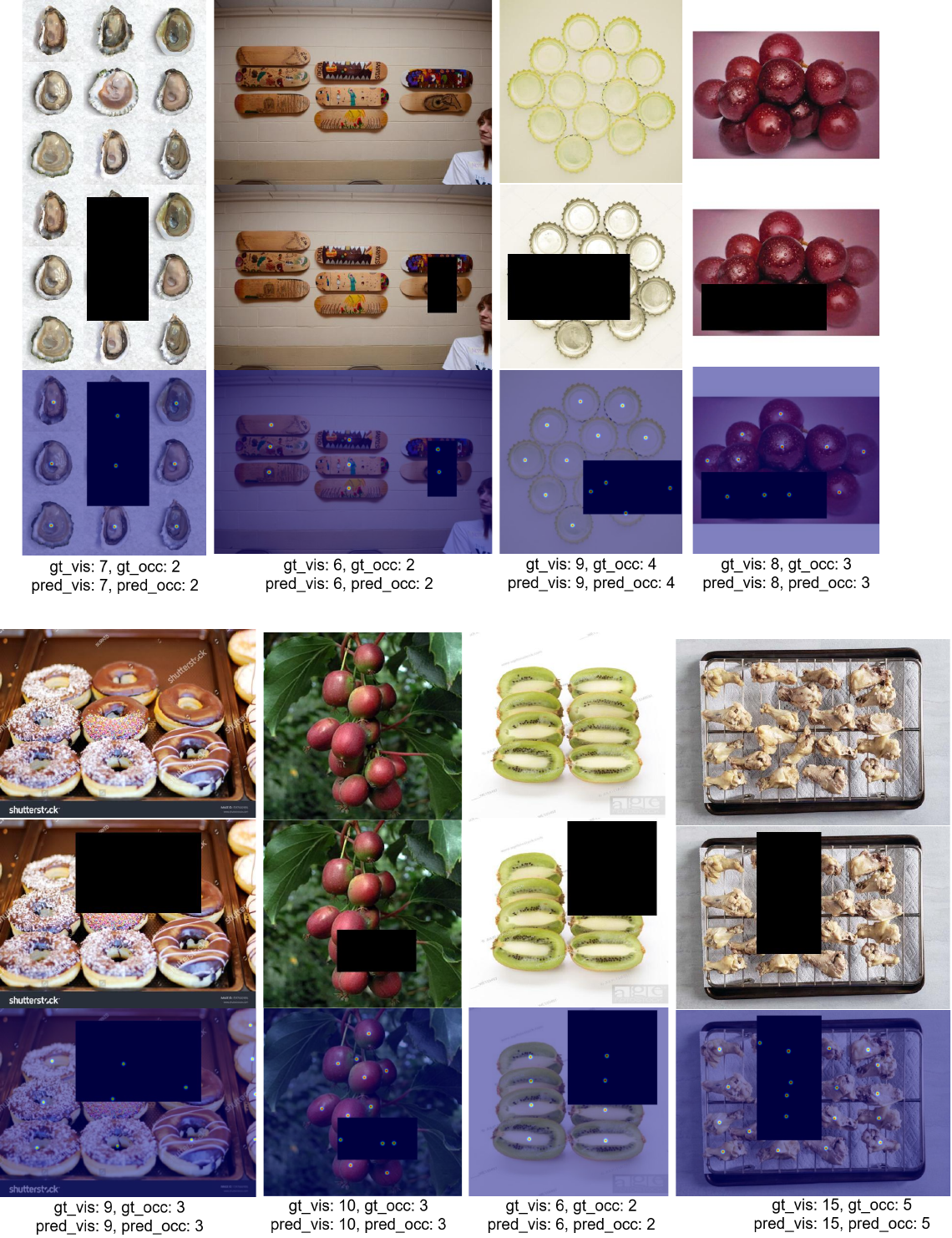}
   \caption{\textbf{Qualitative results on the occluded FSC-147 benchmark.} For each example, we show (top) the original image, (middle) the occluded version with black masks covering objects, and (bottom) our model's predicted density map. The labels indicate ground truth total count ($gt_{total}$), predicted total count ($pred_{total}$), ground truth visible count ($gt_{vis}$), predicted visible count ($pred_{vis}$), ground truth occluded count ($gt_{occ}$), and predicted occluded count ($pred_{occ}$). In these examples, CountOCC correctly estimated both the total count and the breakdown of visible and occluded instances across diverse object categories and scene types.}
   \label{fig:fsc147_qualitative}
\end{figure*}

\begin{figure*}[!htbp]
  \centering
   \includegraphics[width=0.95\linewidth]{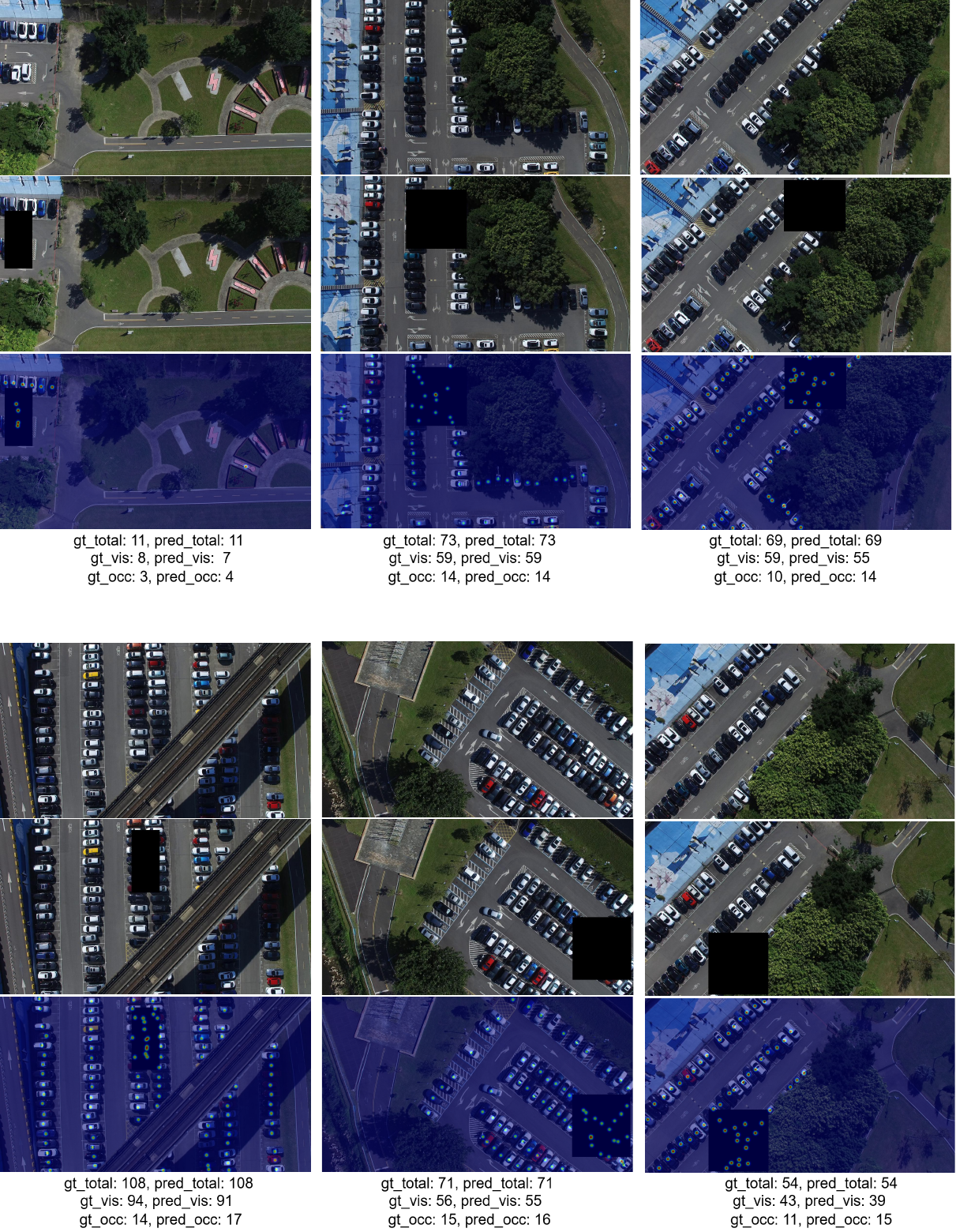}
   \caption{\textbf{Qualitative results on the occluded CARPK benchmark.} For each example, we show (top) the original image, (middle) the occluded version with black masks covering vehicles, and (bottom) our model's predicted density map. The labels indicate ground truth total count ($gt_{total}$), predicted total count ($pred_{total}$), ground truth visible count ($gt_{vis}$), predicted visible count ($pred_{vis}$), ground truth occluded count ($gt_{occ}$), and predicted occluded count ($pred_{occ}$). CountOCC demonstrates robust counting performance on aerial parking lot imagery, accurately estimating total counts despite challenging occlusion patterns and demonstrating strong cross-dataset generalization.}
   \label{fig:carpk_qualitative}
\end{figure*}

\begin{figure*}[!htbp]
  \centering
   \includegraphics[width=0.95\linewidth]{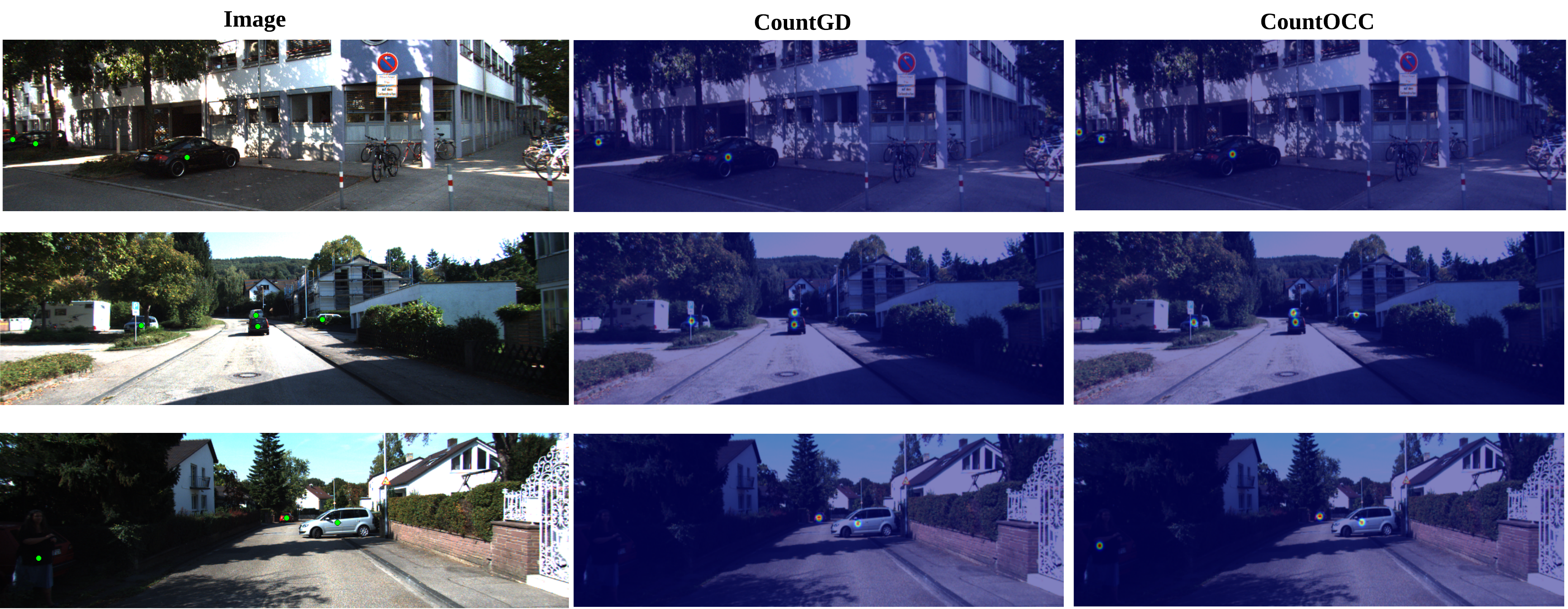}
   \caption{\textbf{Qualitative comparison on the KINS dataset (natural occlusion).} For each example, we show (left) the original image with ground-truth target annotations in \textcolor{bottlegreen}{green}, (middle) the predictions from the CountGD architecture, and (right) our model's predicted density map. In these examples, CountOCC produces total-count estimates closer to the ground truth than CountGD under natural inter-object occlusion.}
   \label{fig:kins_qualitative}
\end{figure*}

\begin{figure*}[!htbp]
  \centering
   \includegraphics[width=0.95\linewidth]{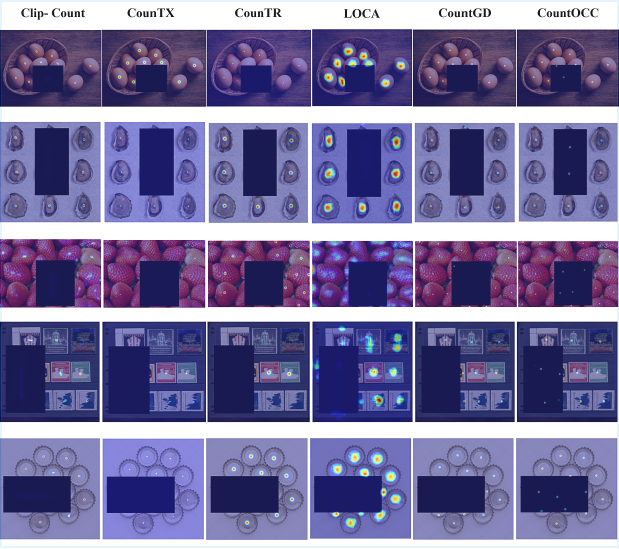}
   \caption{Qualitative comparison on occluded FSC-147. Each column shows predictions from CLIP-Count, CounTX, CounTR, LOCA, CountGD, and CountOCC (ours). Previous counting methods tend to undercount hidden objects, whereas CountOCC produces estimates closer to the ground-truth counts in these examples.}
   \label{fig:fsc147_qualitative_comparison}
\end{figure*}

\begin{figure*}[!htbp]
  \centering
   \includegraphics[width=0.95\linewidth]{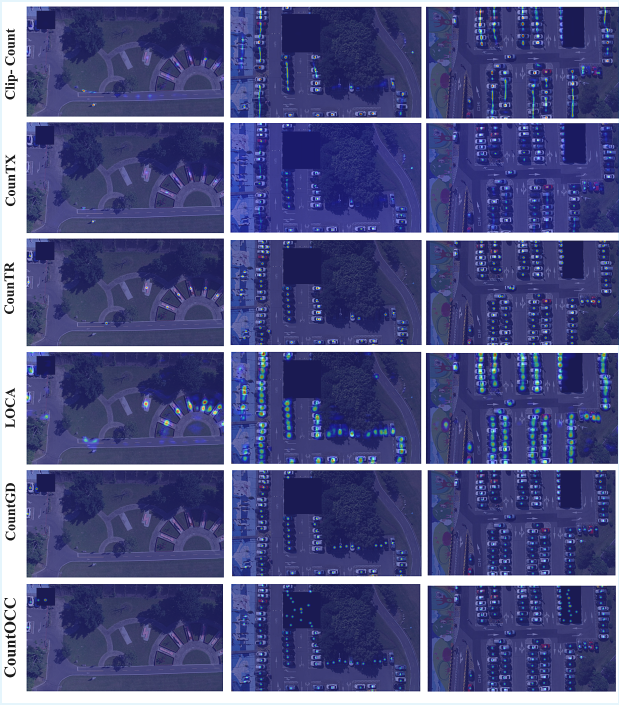}
   \caption{Qualitative comparison on the occluded CARPK benchmark. Rows list predictions from CLIP-Count, CounTX, CounTR, LOCA, CountGD, and CountOCC (ours). Under partial or heavy occlusion, prior methods tend to undercount, whereas CountOCC produces a close match to ground truth across diverse parking layouts.}
   \label{fig:carpk_qualitative_comparison}
\end{figure*}


\section{Visible vs.\ Occluded Performance Breakdown}
\label{sec:performance-visible-occluded}

To better understand behavior under occlusion, \cref{tab:visible_occluded} reports MAE and RMSE separately for visible and occluded instances on FSC-147-OCC.
This per-region evaluation provides a finer-grained view than overall error alone, isolating performance on targets that are explicitly masked versus those that remain visible. CountOCC substantially reduces error on occluded instances compared to CountGD, while largely preserving performance on visible instances across both validation and test splits. Overall, these results suggest that our occlusion-oriented design improves robustness where occlusion is present, while largely preserving performance on visible instances.

\begin{table}[tb]
\centering
\footnotesize
\setlength{\tabcolsep}{2pt}
\caption{Visible-occluded breakdown on FSC-147-OCC.
MAE/RMSE are reported separately for visible and occluded instances within the same images on the validation and test splits.}
\begin{tabular}{lcccccccc}
\toprule
Method &
\multicolumn{4}{c|}{Validation} &
\multicolumn{4}{c}{Test} \\
\cmidrule(lr){2-5}\cmidrule(lr){6-9}
& \multicolumn{2}{c}{Visible} & \multicolumn{2}{c}{Occluded}
& \multicolumn{2}{c}{Visible} & \multicolumn{2}{c}{Occluded} \\
& MAE & RMSE & MAE & RMSE & MAE & RMSE & MAE & RMSE \\
\midrule
CountGD \cite{amini2024countgd}  & 8.05 & \textbf{27.04} & 17.46 & 37.71 & 9.74 & \textbf{53.94} & 18.16 & 48.80 \\
CountOCC & \textbf{8.04} & 30.03 & \textbf{5.61} & \textbf{17.21} & \textbf{8.52} & 55.28 & \textbf{5.80} & \textbf{30.84} \\
\bottomrule
\end{tabular}

\label{tab:visible_occluded}
\end{table}

\section{Performance on Clean Dataset}
\label{sec:performance-clean}
\cref{tab:fsc147_clean_results} reports results on the original (unoccluded) FSC-147 benchmark, using the same evaluation protocol as prior work.
Although CountOCC is designed for amodal counting under occlusion, it remains highly competitive on fully visible images. It achieves accuracy that is close to the best-performing method, CountGD \cite{amini2024countgd}, and consistently improves over text-only and exemplar-only baselines.

We observe a regression compared to CountGD on FSC-147, which suggests a trade-off between occlusion robustness and performance in the fully visible setting. Nevertheless, the results show that CountOCC largely preserves the core strengths of the underlying open-world counting paradigm on clean images while substantially improving robustness under occlusion.

\begin{table}[!htbp]
\centering
\small
\setlength{\tabcolsep}{2pt}

\caption{Performance on the original (unoccluded) FSC-147 benchmark. \textbf{Bold} indicates the best result and \underline{underline} indicates the second best.  $^\dagger$VE: Visual Exemplars.}

\begin{tabular}{lccccc}
\toprule
Method &
Input &
\multicolumn{2}{c}{Validation} &
\multicolumn{2}{c}{Test} \\
& & MAE $\downarrow$ & RMSE $\downarrow$ & MAE $\downarrow$ & RMSE $\downarrow$ \\
\midrule
Patch-selection \cite{xu2023zero}     & Text & 26.93 & 88.63 & 22.09 & 115.17 \\
CLIP-Count \cite{jiang2023clip}          & Text & 18.79 & 61.18 & 17.78 & 106.62 \\
VLCounter \cite{kang2024vlcounter}            & Text & 18.06 & 65.13 & 17.05 & 106.16 \\
CounTX  \cite{AminiNaieni23}                  & Text & 17.10 & 65.61 & 15.69 & 106.06 \\

GroundingREC \cite{dai2024referring}          & Text & 10.06 & 58.62 & 10.12 & 107.19 \\
{CountGD}$_{\mathrm{txt}}$ \cite{amini2024countgd} & Text & 12.14 & 47.51 & 12.98 & 98.35 \\

CounTR \cite{liu2022countr} & VE & 13.13 & 49.83 & 11.95 & 91.23 \\
LOCA \cite{Dukic_2023_ICCV}   & VE & 10.24 & 32.56 & 10.79 & 56.97 \\
DAVE \cite{pelhan2024dave} & VE &  8.91 & 28.08 &  8.66 & 32.36 \\

CountGD \cite{amini2024countgd} & VE \& Text & \textbf{7.10} & \textbf{26.08} & \textbf{5.74} & \textbf{24.09} \\
CountOCC & VE \& Text & \underline{7.24} & \underline{27.10} & \underline{7.02} & \underline{28.76} \\
\bottomrule
\end{tabular}
\label{tab:fsc147_clean_results}
\end{table}

\section{Reconstruction Loss Design}
To validate our reconstruction objective, we ablate the contribution of each loss term on FSC-147-OCC. As shown in \cref{tab:ablation_losses}, using $\mathcal{L}_{\text{$\ell_2$}}$ provides a baseline for feature alignment. Adding $\mathcal{L}_{\text{cosine}}$ yields a clear improvement, reducing validation MAE and RMSE by 12.25\% and 37.87\% over the baseline, and modestly improving test RMSE. Incorporating $\mathcal{L}_{\text{charb}}$ further strengthens validation performance, achieving 18.44\% and 38.83\% reductions in validation MAE and RMSE relative to $\mathcal{L}_{\text{$\ell_2$}}$, though this gain does not translate to improved test RMSE. Finally, the full objective delivers the most consistent performance across splits, reducing validation MAE/RMSE by 16.35\%/54.98\% and test MAE/RMSE by 13.22\%/55.00\% relative to the $\mathcal{L}_{\text{$\ell_2$}}$ baseline. These results indicate that combining complementary supervision terms is important for reliable feature recovery.
\begin{table}[!htbp]
\centering
\small
\caption{Loss-design ablation for feature reconstruction module.}
\label{tab:ablation_losses}
\setlength{\tabcolsep}{3pt}
\begin{tabular}{l c c c c}
\toprule
Experiment & \multicolumn{2}{c}{Validation} & \multicolumn{2}{c}{Test} \\
\cmidrule(lr){2-3} \cmidrule(lr){4-5}
 & MAE & RMSE & MAE & RMSE \\
\midrule
$\mathcal{L}_{\text{$\ell_2$}}$                   &  13.88&78.67  & 13.24 & 88.93 \\
$\mathcal{L}_{\text{$\ell_2$}}$ + $\mathcal{L}_{\text{cosine}}$                 &  12.18&48.88	& 12.38 &	87.04  \\
$\mathcal{L}_{\text{$\ell_2$}}$ + $\mathcal{L}_{\text{cosine}}$ + $\mathcal{L}_{\text{charb}}$      &  \textbf{11.32}&48.12 & 11.90&91.45  \\
$\mathcal{L}_{\text{$\ell_2$}}$ + $\mathcal{L}_{\text{cosine}}$ + $\mathcal{L}_{\text{charb}}$ + $\mathcal{L}_{\text{sim}}$ +  $\mathcal{L}_{\text{cst}}$       &  11.61&\textbf{35.42} & \textbf{11.49}&\textbf{40.02}  \\

\bottomrule
\end{tabular}
\end{table}

\section{Qualitative Results}
\label{sec:qualitative results}

In this section, we present additional qualitative results that illustrate how CountOCC behaves under challenging occlusion patterns on both FSC-147-OCC and CARPK-OCC. 

\cref{fig:fsc147_qualitative} visualizes representative examples from the occluded FSC-147 benchmark across diverse object categories and scene layouts. For each image, we show the original scene, its occluded counterpart with black masks, and the predicted density map, together with the ground-truth and predicted counts for the total, visible, and occluded instances. These examples highlight that, even when a large fraction of the target objects is fully covered, CountOCC produces spatially coherent density maps and total-count estimates that remain close to the ground truth, together with consistent visible and occluded count estimates.

\cref{fig:carpk_qualitative} shows analogous visualizations on the occluded CARPK-OCC benchmark, focusing on aerial parking-lot scenes with structured layouts and frequent inter-object overlap. Here, CountOCC robustly localizes vehicles in dense configurations, preserves sharp density responses over visible cars, and infers plausible contributions for masked regions. Across both datasets, these qualitative results complement our quantitative evaluations by visually demonstrating that the model not only matches the ground-truth totals but also maintains consistent amodal reasoning about how many instances are visible versus hidden behind occluders.


\section{Qualitative Comparison to Other Methods}
\label{sec:qualitative comparison}

In this section, we provide qualitative comparisons between CountOCC and prior open-world counting methods on KINS, FSC-147-OCC, and CARPK-OCC.

\cref{fig:kins_qualitative} presents the qualitative comparison on the KINS dataset, where occlusions occur naturally in real street-scene images. Compared with CountGD, CountOCC produces more consistent density responses around target objects under inter-object overlap and partial visibility, leading to more reliable total-count estimates. These results further show that the proposed approach generalizes beyond synthetic occlusion patterns and remains effective in realistic occluded scenes.

\cref{fig:fsc147_qualitative_comparison} shows representative results on FSC-147-OCC under varying degrees of occlusion. Existing methods (CLIP-Count \cite{jiang2023clip}, CounTX \cite{AminiNaieni23}, CounTR \cite{liu2022countr}, LOCA \cite{Dukic_2023_ICCV}, and CountGD \cite{amini2024countgd}) generally capture visible instances but exhibit a strong bias toward foreground evidence, leading to undercounting when a substantial portion of the objects is partially or fully hidden. In contrast, CountOCC produces density maps that more closely align with the ground-truth totals, allocating meaningful density mass to occluded regions and demonstrating effective amodal feature reconstruction across diverse object categories and scene layouts.

\cref{fig:carpk_qualitative_comparison} presents analogous comparisons on CARPK-OCC, where we synthetically occlude vehicles in aerial parking-lot images. As the occlusion level increases, baseline methods progressively lose density in masked regions and underestimate the true counts. CountOCC, however, maintains close agreement with the ground-truth totals across different occlusion patterns, indicating that it can reliably infer the presence of masked vehicles. Together, these results highlight the robustness of our approach for open-world amodal counting on both unstructured and structured scenes.



\section{Failure Cases and Ambiguous Occlusion}
\label{sec:failure}

\begin{figure*}[!htbp]
    \centering
    \includegraphics[
        width=0.95\textwidth
    ]{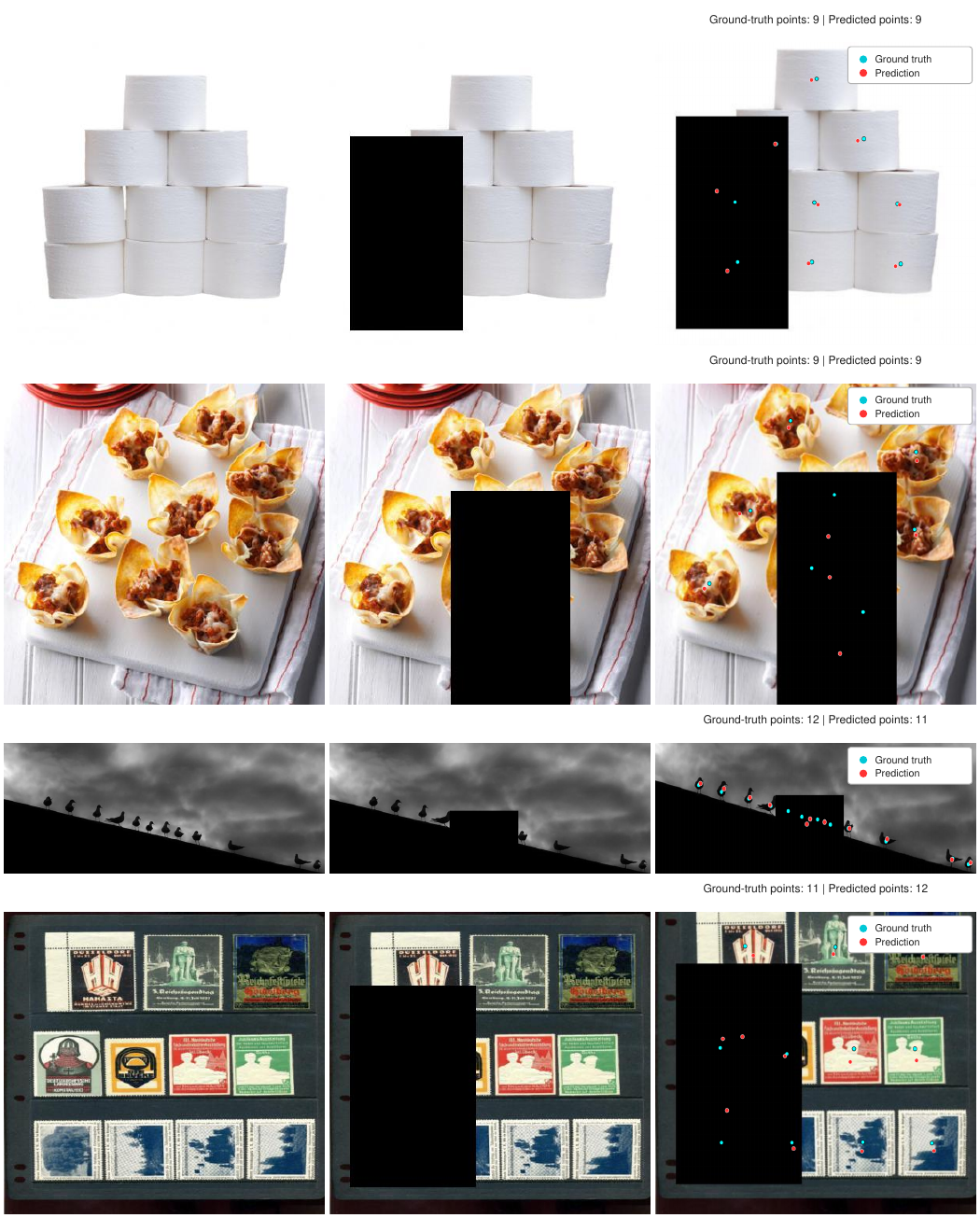}
    \caption{
Failure cases and ambiguous occlusion on FSC-147-OCC. Each row shows the original image, the corresponding occluded input, and the predicted points overlaid with ground-truth points. The first two rows match the annotated hidden counts, whereas the seagull and stamp examples show under- and over-counting, respectively. These cases expose an inherent limitation: when the visible context admits multiple plausible hidden configurations, CountOCC cannot uniquely determine the number of fully hidden instances and instead estimates the most plausible count from the available evidence.
}
    \label{fig:qualitative_ambiguity}
\end{figure*}

To examine ambiguity in fully hidden-instance counting, Fig.~\ref{fig:qualitative_ambiguity} presents four representative cases from FSC-147-OCC. Each row shows the original image, the corresponding occluded input, and the predicted points overlaid with ground-truth points. In the stacked toilet-paper scene, regular spacing provides strong continuation cues and the hidden count is estimated correctly. In the irregular food scene, the annotated count is also obtained despite the absence of a unique visual continuation. In the seagull case, CountOCC predicts three hidden birds when four are annotated, whereas in the stamp case, it predicts five hidden stamps when four are annotated. These examples emphasize that CountOCC estimates a conditional count from visible spatial context and learned regularities, rather than recovering unobservable content from ambiguous occlusion.

\section{Limitations}
\label{sec:limitations}

CountOCC is designed to recover counting-relevant evidence under occlusion, and its effectiveness naturally depends on the quality of the occlusion support used to guide feature reconstruction. When the predicted support closely matches the true occluded regions, the model can reliably reconstruct missing evidence and improve amodal count estimation. In more challenging cases, inaccurate or overly broad masks may introduce additional uncertainty within the reconstructed region, occasionally leading to spurious density responses. Without an occlusion mask, CountOCC remains usable as a standard open-world counter over visible evidence, but explicit recovery of fully occluded instances requires an estimated occlusion support. Improving mask estimation and jointly learning occlusion support with amodal counting are promising directions for future work.


\section{Broader Impacts}
\label{sec:broader impacts}

Amodal counting, the ability to infer object counts even when instances are partially or fully occluded, has the potential to benefit a wide range of real-world applications where visibility is intrinsically limited. Examples include crop and yield estimation from aerial imagery in agriculture, inventory tracking in densely packed warehouses, cell and organism quantification in biological imaging, and robust perception for robots operating in cluttered or unstructured environments. By enabling models to reason beyond directly visible evidence, such systems can serve as more reliable components in downstream decision-making pipelines.

At the same time, the capability to estimate the presence of hidden entities raises important ethical and societal considerations. In particular, applications in surveillance, privacy-sensitive settings, or military contexts could leverage amodal counting in ways that may conflict with expectations of privacy, autonomy, or safety. We therefore emphasize that any deployment of our approach should be accompanied by careful, context-dependent assessment of risks, adherence to relevant regulations, and oversight mechanisms that prioritize responsible use and societal benefit.

\end{document}